# MULTILINGUAL MULTIWORD EXPRESSIONS

M.Tech Report

Submitted in partial fulfillment of the requirements

for the degree of

**Master of Technology**

by

**Lahari Poddar**

Roll No:113050029

Under the guidance of

**Prof. Pushpak Bhattacharyya**

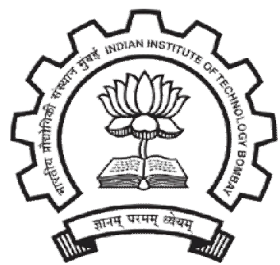

Department of Computer Science and Engineering

Indian Institute of Technology, Bombay

Mumbai

# Dissertation Approval Certificate

Department of Computer Science and Engineering

Indian Institute of Technology Bombay

The dissertation titled 'Multilingual Multiword Expressions' submitted by Lahari Poddar (Roll No: 113050029) is approved for the degree of Master of Technology in Computer Science and Engineering from Indian Institute of Technology Bombay.

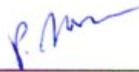

Prof. Pushpak Bhattacharyya
CSE Dept, IIT Bombay
Supervisor

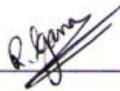

Dr. Ganesh Ramkrishnan
CSE Dept, IIT Bombay
Internal Examiner

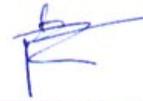

Dr. Ananthakrishnan Ramanathan
Reliance Industries Limited
External Examiner

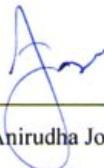

Prof. Anirudha Joshi
IDC Dept, IIT Bombay
Chairperson

Place: IIT Bombay
Date: 18.06.2013

# Declaration

I declare that this written submission represents my ideas in my own words and where others' ideas or words have been included, I have adequately cited and referenced the original sources. I also declare that I adhere to all principles of academic honesty and integrity and have not misinterpreted or fabricated any idea/data/fact/source in my submission. I understand that any violation of the above will be cause for disciplinary action by the Institute and can also evoke penal action from the sources which have not been properly cited or from whom proper permission has not been taken when needed.

*Lahari Poddar*
Signature

LAHARI PODDAR
Name of the student

113050029
Roll Number

18.06.2013
Date

# Abstract

The project aims to provide a semi-supervised approach to identify Multiword Expressions in a multilingual context consisting of English and most of the major Indian languages. Multiword expressions are a group of words which refers to some conventional or regional way of saying things. If they are literally translated from one language to another the expression will lose its inherent meaning.

Example:

*English:* apple of an eye

*Hindi:* आँखों का तारा *(Aankhon ka taara)*

*Bengali:* চোখের মণি *(chokher moni)*

To automatically extract multiword expressions from a corpus, an extraction pipeline have been constructed which consist of a combination of rule based and statistical approaches. There are several types of multiword expressions which differ from each other widely by construction. We employ different methods to detect different types of multiword expressions. Given a POS tagged corpus in English or any Indian language the system initially applies some regular expression filters to narrow down the search space to certain patterns (like, reduplication, partial reduplication, compound nouns, compound verbs, conjunct verbs etc.). The word sequences matching the required pattern are subjected to a series of linguistic tests which include verb filtering, named entity filtering and hyphenation filtering test to exclude false positives. The candidates are then checked for semantic relationships among themselves (using Wordnet). In order to detect partial reduplication we make use of Wordnet as a lexical database as well as a tool for lemmatizing. We detect complex predicates by investigating the features of the constituent words. Statistical methods are applied to detect collocations. Finally, lexicographers examine the list of automatically extracted candidates to validate whether they are true multiword expressions or not and add them to the multiword dictionary accordingly. A universal web service has been developed in order to facilitate multiword expression extraction across the various research groups in India.

Apart from Multiword Expressions Extraction, in this report a Common Concept Hierarchy is also proposed which is a linked structure of wordnets of 18 different Indian languages, Universal Word dictionary and the Suggested Upper Merged Ontology (SUMO). The system is encoded in Lexical Markup Framework (LMF) and we propose modifications in LMF to accommodate Universal Word Dictionary and SUMO. This standardized version of lexical knowledge base of Indian Languages can now easily be linked to similar global resources.

# Acknowledgement

I would like to take this opportunity to sincerely thank Prof. Pushpak Bhattacharyya for his insights, constant support and encouragement. His guidance has been my primary source of motivation.

I would like to thank Munish Minia for making his report and work available for reference. I am grateful to Manish Shrivastava, Dhirendra Singh, Kashiviswanatha Sarma, Samir Janardan Sohoni for their valuable insights and intellectual contribution towards the project. I am thankful to all the members of CFILT at IIT Bombay for their keen interest and contributions, direct or indirect towards the project. It is the detailed discussions and brainstorming analysis carried out at our weekly meetings that has kept me motivated and deeply interested in this topic. I would finally like to thank my parents and friends who always had faith in me and motivated me to make this possible.

# Table of Contents

# List of Figures

# Chapter 1

# Introduction

Lexemes or tokens are basic units of natural language. Following the syntactic structure of a natural language they come together and interact with each other to form a meaningful sentence. Sometimes they convey a meaning as a single unit and sometimes multiple simple words function as a single lexical unit to convey a meaning.

## 1.1. Introduction to Multiword Expressions

Multi Word Expressions are defined as an expression crossing word boundaries that refer to some conventional or regional way of saying things[19]. They are arbitrary word combinations that are very frequent in natural language hence they are also termed as collocations. In Wordnet 1.7 around 41% of the entries are multiword. Informally, multiword refers to a group of words, if literally translated from one language to another will lose their inherent meaning.

As an example of collocation, let us consider the term *strong coffee*. We always use the adjective *strong* to describe *coffee* or *tea* but not its other synonyms (say, *powerful*). Whereas in case of *'drugs'*, *'powerful drug'* is more frequently used than *'strong drug'*. There is no specific reason as to why one representation or interpretation should be chosen over the other. It is just the way it is. Multiwords are random, arbitrary and idiosyncratic. They are completely language dependent and are transparent only to a native speaker of the language.

Examples:
1. Pitter patter drops the rain

    [Hindi] टिप टिप पानी बरस रहा है (Tip tip paani baras raha hain)

    [Bengali] টাপুর টুপুর বৃষ্টি পরে (tapur tupur bristi pore)

2. He is working day and night

    [Hindi] वोह दिन रात काम कर रहा हैं (woh din raat kam kar raha hain)

[Bengali] ও সারা দিন রাত কাজ করছে (o sara din raat kaj korche)

3. [Hindi] घर घर में दीप जले (ghar ghar mein deep jale)

   [Bengali] ঘরে ঘরে দীপ জ্বলে (ghore ghore deep jale)

   Gloss: home home candle lit

   Translation: In every home candle is lit

4. Down to earth person

   [Bengali] মাটির মানুষ (matir manush)

## 1.1. Motivation

The motivation behind extracting collocations from text is that this collocational information will be useful in a number of Natural Language Processing tasks.

- **Machine Translation**: Collocations differ from language to language. So a multiword expression cannot directly be translated from one language to another conserving its inherent idiosyncrasy or metaphoric meaning.

  For example: In Hindi *'aankhon ka taara'* means someone's favorite person. The same meaning is conveyed by the English idiom *'apple of an eye'*. But they are not direct literal translations of one another. As another example, we have the expression *red handed* in English which means to catch someone while in the act of committing a crime has its Hindi counterpart as *'range haath'* (not 'laal haath', the literal translation).

  Following is a snapshot from Google translator as an evidence of the fact that an automatic machine translation system needs to have the knowledge of multiword expressions of a language. As we can see in **Figure 1.1** the translations done by a machine translation system from English to Hindi is erroneous due to the ignorance of multiword expressions in the sentence.

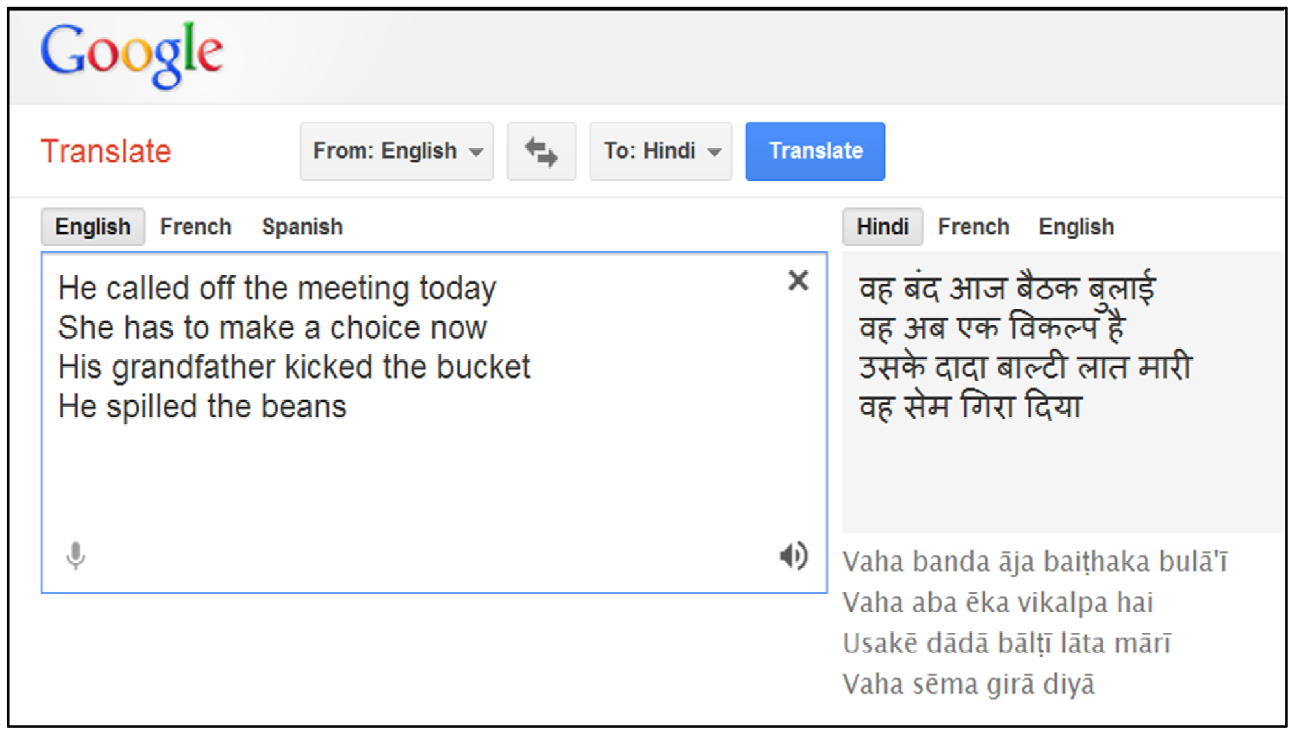

**Figure 1.1:** Machine translation of MWEs

- **Natural Language Generation**: Natural Language Generation refers to the act of generating text in natural language from a logical symbolic form. This task requires appreciating the nuances of the language. Collocation is one of them. If collocations are not accounted for while generating the text then some word combinations may accidentally occur in the text which have some inline meaning. This ignorance will affect the linguistic quality of the generated text.

- **Automatic Simplification of Text:** Some of the text editors allow automatic simplification or modification text by providing synonyms of a particular word or expression. If it doesn't have the knowledge of collocations then it might suggest synonyms which are inappropriate or while simplifying an expression it might lose its metaphorical meaning. So such tools need to be aware of the collocational constraints of the language.

- **Enhancing natural language lexical resources:** Multiword expressions give crucial knowledge about a language. They are highly prevalent and very irregular in nature. Hence MWEs must be stored in lexicons of natural language processing applications like Wordnet and disambiguated universal word dictionaries.

    For example:

- उसके दादाजी ने **दम तोडा**
- Transliteration: uske dadaji ne dam toda
- Gloss: His grandfather died.
- Translation: His grandfather **kicked the bucket**.

In the above example *kick the bucket* and *दम तोडा* means *to die*. So these multiword expressions should enter the corresponding synsets in wordnet as well as universal word dictionary.

Motivated by the importance of multiword expressions in natural language processing tasks we, at CFILT IIT Bombay chose to address the problem of extracting multiword expressions from a given text corpora. We have developed a pipeline for extracting multiword expressions and making a repository of these expressions so that other fields of NLP can benefit from it. The current work contributes to the Cross Lingual Information Access (CLIA) and Indian Language Machine Translation efforts being undertaken by consortia of academic institutions across India.

## 1.2. Organization of the report

In this chapter, we have introduced the concept of MWEs and explained the various fields of Natural Language Processing that needs to be concerned about the knowledge of Multiword Expressions. In Chapter 2, at first the MWEs are formally defined. We also present a classification of MWEs according to different criteria and the necessary and sufficient condition for an expression to be classified as MWE. In Chapter 3 we describe the different methods to extract MWEs from a corpus and class them as statistical method and knowledge-based method. In this chapter, another contemporary ongoing project in this research field for extracting MWEs is described. Chapter 4 describes my contribution in the development of MWE Engine in IIT Bombay. Chapter 5 describes some experiments that I have performed and the ongoing research activity regarding the project at IIT Bombay. Chapter 6 gives the evaluation of the performance of the system. In Chapter 7 we present another contribution of ours, a Common Concept Hierarchy which is built by merging wordnet, universal word dictionary and an upper ontology. Chapter 8 describes the construction of a generic stemmer, its functionalities and applications. Finally, Chapter 9 concludes this report with our perspectives and listing our future goals.

# Chapter 2

# Background

In this section we will describe the formal definition Multiword Expression along with the necessary and sufficient conditions for an expression to be termed as MWE. The different types and characteristics possessed by such expressions are elaborated. We will also specify the necessary and sufficient conditions for an expression to be classified as an MWE. Finally we will look at the different tasks to be done for extracting MWEs from a text.

Various researchers have defined multiword expressions differently during their research. We'll present some of the definitions here and it can be observed that all of them primarily refer to a single central concept.

- A collocation is an expression consisting of two or more words that correspond to some conventional way of saying things.[19]
- Idiosyncratic interpretations that cross word boundaries (or spaces) [18]
- Recurrent combinations of words that co-occur more frequently than chance, often with non-compositional meaning[20]
- A pair of words is considered to be a collocation if one of the words significantly prefers a particular lexical realization of the concept the other represents[14]

## 2.1. Features of Multiword Expressions

There are certain features that a group of words must have in order to be treated as collocation. The principal features are:

- **Non-Compositionality:** The meaning of a complete multiword expression can't completely be determined from the meaning of its constituent words.

    The meaning of the expression might be completely different from its constituents (the idiom *kick the bucket* means *to die*) or there might be some added element or inline meaning to it that cannot be predicted from the parts(the phrase *back to square one* means to reach back to the place from where one had started).

- **Non-Substitutability:** The components of a multiword expression cannot be substituted by one of its synonyms without distorting the meaning of the expression even though they refer to the same concept.

    For example, in the expression *bread and butter* the component words cannot be replaced by their synonym keeping the meaning(to earn one's daily living) intact.

- **Non-Modifiability:** Many collocations cannot be freely modified by grammatical transformations (like, change of tense, change in number, addition of adjective etc.). These collocations are frozen expressions, they cannot be modified in any way.

    For example, the idiom *let the cat out of the bag* cannot be modified to \**let the big cat out of the bag* or something similar.

## 2.2. Types of MWEs

Collocations or Multiword Expressions can be classified into different classes according to their lexical and semantic characteristics. The classification as described in [18] is given below.

1) **Lexicalized Phrases:** This type of phrases have some form of idiosyncratic or added meaning to the structure. They are either syntactically idiosyncratic or semantically non-decomposable. Lexicalized phrases can be classified into 3 parts.

a) **Fixed Expressions:** This is the class of expressions that defy the general conventions of grammar and compositional interpretations. These expressions are completely frozen and do not undergo any modifications at all.

    Example: in short, of course, ad hoc, by and large

b) **Semi-Fixed Expressions:** This type of expressions have restrictions on word order and the structure of the phrase but they might undergo some form of lexical variations. Semi-Fixed expressions can be further classified into 3 subtypes:

(1) **Non-Decomposable Idioms:** Depending on their semantic composition, idioms can be classified into two types: Decomposable and Non-Decomposable.

For decomposable idioms each component of the idiom can be assigned a meaning related to the overall meaning of the expression. For the idiom *spill the beans*, 'spill' can be assigned the sense of 'reveal' and 'beans' can denote the sense of 'secret'. But in case of Non-Decomposable idioms no such analysis is possible.

For the idiom *kick the bucket* none of its components can be assigned a sense such that the overall idiom means 'to die'.

It is these Non-Decomposable idioms which are semi-fixed. Due to their opaque meaning they do not undergo any syntactic variations but might allow some minor lexical modification (*kick the bucket -> kicked the bucket*).

(2) **Compound Nominals:** Compound nominals also do not undergo syntactic modifications but allow lexical inflections for number i.e. they can be changed to their singular or plural form.

Example: car park, part of speech, railway station

(3) **Named Entities:** These are syntactically highly idiosyncratic. These entities are formed based on generally a place or a person.

Example: the cricket team names in IPL are formed based on the region. In a proper context the team names are often mentioned without the name of the place, like '(Kolkata) Knight Riders', 'Royal Challengers (Bangalore)' etc. When the team name occurs as a modifier in some compound noun a modifier is added ('the Kolkata Knight Riders player...' )

c) **Syntactically-Flexible Expressions:** As opposed to the strict word order constraint of Semi-Fixed expressions, Syntactically-Flexible expressions allow a wide variety of syntactic variations. They can be classified into 3 types:

(1) **Verb-Particle Construction:** Verb-Particle constructions or phrasal verbs consist of a main verb and a particle. Transitive verb-particle constructions are a good example of non adjacent collocations as they can take an NP argument in between (like, *call him up*).

Example: call off, write up, eat up etc.

(2) **Decomposable idioms:** Decomposable idioms are syntactically flexible and behave like semantically linked parts. But it's difficult to predict exactly what type of syntactic variations they undergo.

Example: spill the beans, let the cat out of the bag

(3) **Light-Verb Constructions:** Verbs with little semantic content (make, take, do) are called light verbs as they can form highly idiosyncratic constructions with some nouns.

Example: *make a decision , do a favor, take a picture etc* are light-verb constructions as there is no particular reason why *do me a favor* should be preferred over \**make me a favor* and so on.

2) **Institutionalized Phrases:** These phrases are completely compositional (both syntactically and semantically) but are statistically idiosyncratic. These are just fixed terms which do not have any alternate representations.

Example: traffic light, fresh air, many thanks, strong coffee etc.

## 2.3. Classification of MWEs

Multiword Expressions can be classified into the following groups:

1. **Reduplication:** Reduplication is a morphological process by which the root or stem of a word, or part of it, is repeated[29]
    i. **Onomatopoeic Reduplication:** The constituent words do not have any dictionary meaning; rather they imitate a sound or an action along with the sound.

Example:

1. knock knock

    Gloss: The sound of knocking at door

2. টিক টিক (Hindi)

    Transliteration: tik tik

    Gloss: The sound of a clock

    Translation: tik tik

3. खट खट (Hindi)

    Transliteration: khat khat

    Gloss: The sound of knocking at door

    Translation: knock knock

4. ছম ছম (Bengali)

    Transliteration: chham chham

    Gloss: the sound of anklets

5. பட பட (Tamil)

    Transliteration: pada pada

    Translation: fluttering of wings

ii. **Non-Onomatopoeic Reduplication:** The constituent words are meaningful and they are repeated to convey some particular sense.

Example:

1. slowly slowly

2. अहिस्ता अहिस्ता (Hindi)

    Transliteration: ahista ahista

    Gloss: slowly slowly

3. চলতে চলতে (Bengali)

    Transliteration: cholte cholte

    Gloss: walking walking

    Translation: while walking

4. வழ வழ (Tamil)

      Transliteration: vazha vazha

      Gloss: smooth smooth

      Translation: Very smooth texture

   5. வீட்டுக்குவீடு (Tamil)

      Transliteration: veetukku veedu

      Gloss: to house house

      Translation: Every house

2. **Partial Reduplication:** Only one of the words among the constituent words is meaningful while the second word is constructed by partially reduplicating the first word.

   Example:

   1. আবোল তাবোল (Bengali)

      Transliteration: abol tabol

      Translation: gibberish

   2. বোকা সোকা (Bengali)

      Transliteration: boka soka

      Translation: foolish

   3. पानी वाणी (Hindi)

      Transliteration: pani vani

      Translation: water

   4. கோவில் குளம் (Tamil)

      Transliteration: kovil kulam

      Gloss: temple pond

      Translation: temple

3. **Semantic Relationship:** Sometimes the paired words have some semantic relationship among themselves

   Example:

   **Synonym**

1. धन दौलत (Hindi)

    Transliteration: dhan daulat

    Translation: wealth

2. সুখ শান্তি (Bengali)

    Transliteration: sukh shanty

    Gloss: Happiness and peace

**Antonym**

3. दिन रात (Hindi)

    Transliteration: din raat

    Gloss: day and night

    Translation: round the clock

4. জীবন মরণ (Bengali)

    Transliteration: jibon moron

    Gloss: life and death

5. ராத்திரி பகலா (Tamil)

    Transliteration: rathiri pagala

    Translation: all the time

    Gloss: night and day

**Sister Words**

6. चाये पानी (Hindi)

    Transliteration: chaye pani

    Gloss: tea water

    Translation: tea and snacks

7. লেখা পড়া (Bengali)

    Transliteration: lekha poda

    Gloss: writing reading

    Translation: study

4. **Collocations:** Collocations are statistical idiosyncrasies of a language. They do not have any syntactical peculiarity or share any semantic relationship. These are just fixed expressions which appear very frequently in natural language without undergoing any modifications. Example: traffic light, bus driver, fresh air

5. **Light Verb Constructs:** Light verb constructs are formed by the combination of Noun + Verb where the verb has lost its meaning partially but the noun is used in its original sense.[25]

   Example: have a seat, make a decision, do a favor, take a picture

   In the above examples as we can see in the noun verb constructs the verbs do not carry much meaning i.e. there is no reason why we should say *'have a seat'* instead of \**'take a seat'*. Due to this eccentricity we need to store such constructs but not all Noun + Verb combinations are light verb constructs. Consider the examples:

   - जम्हाई लेना (Hindi)

     Transliteration: jamhai lena
     Gloss: yawn take
     Translation: To yawn

   - चाय लेना (Hindi)

     Transliteration: chaye lena
     Gloss: tea take
     Translation: To take tea

   In the above examples the first one is a light verb construct (since the verb doesn't convey much meaning) whereas the second one is completely compositional. So it is necessary to detect true light verb constructs and store them instead of storing all Noun+Verb combinations.

6. **Compound Verb:** A compound verb or complex predicate is a multi-word compound that acts as a single verb. One component of the compound is a *light verb* or *vector*, which carries any inflections, indicating tense, mood, or aspect, but provides only fine shades of meaning.

The other, "primary", component is a verb which carries most of the semantics of the compound, and determines its arguments.[28]

Compound verbs are very common in Indian languages. It is represented by Verb + Verb combination where the first verb is called 'polar verb' and the second verb is called 'vector verb'.

Example:

- राम घर से निकल गया (Hindi)

    Transliteration: Ram ghar se nikal gaya

    Gloss: Ram home of out went

    Translation: Ram went out of home

- আমি এই কাজটা করে দেব ( Bengali)

    Transliteration: ami ei kajta kore debo

    Gloss: I this work do will

    Translation: I will do this work

- சென்று விடு (Tamil)

    Transliteration: Senru vidu

    Translation: Go away

Compound verbs i.e. this Verb + Verb combinations are very rare in English but this similar fact can be observed in English with Verb + Preposition combinations.

Example:

- **English:** Finish off the work fast
- **Bengali:** কাজটা তাড়াতাড়ি করে ফেলো (kajta taratari kore felo)
- **Hindi:** य़ेह काम जल्दी ख़तम कर दो (yeh kaam jaldi khatam kar do)

7. **Compound Noun:** Compound noun refers to the phenomena where two or more nouns combine to form a new compound. This type of compounds is highly generative, i.e. new compounds get added to the language very frequently as the language evolves. Noun compounds belong to the most frequent classes of Multiword Expressions.

Example:

Science fiction writer, railway station

- எதிர் கட்சி (Tamil)

    Transliteration: ethir katchi
    Gloss: opposition party

- கண் மை (Tamil)

    Transliteration: kan mai
    Gloss: kajal

## 2.4. Necessary and Sufficient Conditions for MWE

In this section we describe the standard that has been agreed upon during Kashmir Multiword Workshop 2011. The necessary and sufficient conditions for an expression to be classified as MWE is as follow:

### 2.4.1 Necessary Conditions

For a word sequence to be a MWE, it has to be separated by space/delimiter.

Example:

- இந்திய கிரிக்கெட் அணி  (Tamil)

    **Transliteration:** Indhiya kirikket ani
    **Gloss:** India cricket team
    **Translation:** Indian Cricket Team

### 2.4.2 Sufficient Conditions

The sufficient conditions to be an expression to be classified as MWE are:

1. The non-compositionality of meaning of the MWE, i.e. meaning of a MWE cannot be derived from its constituents.

    Examples:

    - చెట్టు కింnదికి ప్లీడరు(Telugu)

        **Transliteration**: cevttu kimda plidaru

**Gloss:** a lawyer sitting under the tree

**Translation:** an idle person

2. The fixity of expression, i.e. the constituents of MWE cannot be replaced by its synonyms or other words.

   Examples:
   - Correct: life imprisonment
     
     *Incorrect: lifelong imprisonment
   - Correct: Many thanks
     
     *Incorrect: Plenty thanks

## 2.5. MWE Extraction tasks

Armed with the background knowledge of the definition and features of Mutiword Expressions we tried to classify them into different classes according to the most suitable extraction approach. **Figure 2.1** shows the classification.

As we can see, both statistical and rule based approaches are necessary to solve this problem. We can also see from the following figure the stack of NLP tools that need to be deployed in order to identify different classes of Multiword Expressions.

| ML \ NLP | String + Morphology | POS tagging | POS tagging + Wordnet | POS tagging + List | Chunking | Parsing |
|---|---|---|---|---|---|---|
| **Rules** | Onomaetopic Reduplication *(tik tik, chham chham)* | Non-Onomaetopic Reduplication *(ghar ghar)* | Semantic relation (Synonym, Antonym, Hypernym) *(raat din, dhan doulat, chaye paani)* | | | Non-contiguous something |

| Statistical | | Colloctions or fixed expressions *(many thanks)* | | Conjunct verb (verbalizer list), Compund verb (verctor verb list) *(salaha dena, has uthama)* | | Non-contiguous Complex Predicate |

**Figure 2.1**: MWE extraction tasks (Pushpak Bhattacharyya, LREC 2012 )

# Chapter 3

# MWE Extraction Approaches

## 3.1. Approaches by various researchers

In this section we are going to present a survey of the different approaches tried out by different researchers over the years in order to extract multiword expressions from a text. The methods vary widely from one another. Some of them have taken a Linguistic approach, some have used statistical techniques and some have taken help of the open source resources available to us to solve the problem.

### 3.1.1 Rule Based Approaches

There have been quite a few approaches which try to detect multiwords by leveraging the rules forming them in the first place.

#### *3.1.1.1.  Identification of Reduplication in Bengali*

Reduplication is a subtype of Multiword Expressions and a method for identifying reduplications and then classifying them has been reported by the authors. Reduplications have been categorized into 2 levels in [5], namely **Expression Level** and **Sense Level.** They can be further subcategorized as:

**Expression Level:**
  a) Onomatopoeic expressions: The constituent words imitate a sound or a sound of an action. Generally in this case the words are repeated twice with the same 'matra'.
   - ঝম ঝম (Bengali)

     Transliteration: jham jham

     Translation: the sound of rain
   - টপ টপ  (Bengali)

     Transliteration: top top

     Translation: the sound of dropping water

b) Complete Reduplication: The constituent words are meaningful and they are repeated to convey some particular sense.

- চলতে চলতে (Bengali)

    Transliteration: chalte chalte

    Gloss: walking walking

    Translation: while walking

- বার বার (Bengali)

    Transliteration: bar bar

    Gloss: time time

    Gloss: time and again/ repeatedly

c) Partial Reduplication: In partial reduplication generally three cases are possible

    (i) change of the first vowel or the matra attached with first consonant

    (ii) change of consonant itself in first position

    (iii) change of both matra and consonant

- বোকা সোকা (Bengali)

    Transliteration: boka soka

    Translation: Foolish

- চাল চুলো (Bengali)

    Transliteration: chal chulo

    Translation: belongings

d) Semantic Reduplication: A dictionary based approach was followed to identify consecutive occurrences of synonyms and antonyms.

- দিন রাত (Bengali)

    Transliteration: *din-raat*

    Gloss: day and night

    Translation: round the clock/ all the time

- পাপ পুণ্য (Bengali)

    Transliteration: paap-punyo

    Gloss: sin and virtue

**Sense Level Classification:**

a) Sense of repetition:

- রোজ রোজ ( Bengali)

    Transliteration: roj roj

    Gloss: day day

    Translation: everyday

- বছর বছর (Bengali)

    Transliteration: bachor bachor

    Gloss: year year

    Translation: every year

b) Sense of plurality:

- ছোটো ছোটো ( Bengali)

    Transliteration: choto choto

    Gloss: small small

    Translation: small

c) Sense of Emphatic :

- সুন্দর সুন্দর (Bengali)

    Transliteration: sundor sundor

    Gloss: beautiful beautiful

    Translation: beautiful

- লাল লাল (Bengali)

    Transliteration: laal laal

    Gloss: red red

    Translation: red

d) Sense of completion :

- খেয়ে দেয়ে (Bengali)

    Transliteration: kheye deye

    Translation: after finishing meal

e) Sense of incompleteness :

- বলতে বলতে (Bengali)

    Transliteration: bolte bolte

    Gloss: talking talking

    Translation: while talking

- চলতে চলতে (Bengali)

    Transliteration: cholte cholte

    Gloss: walking walking

    Translation: while walking

Some collected articles of Rabindranath Tagore have been used as a corpus. The system developed by them reportedly achieved 92% precision and a recall of 91%. There exists some combination of words which have a semantic relationship between them but are not exactly synonyms or antonyms of each other (for eg: '*slow and steady*'). The system was unable to detect such type of reduplications using only a dictionary.

### *3.1.1.2.    Detecting noun compounds and light verb constructions*

The authors have described some rule based methods to detect noun compounds and light verb constructions in running texts [25].

Noun compounds are productive, i.e. new nominal compounds are being formed in language use all the time, which yields that they cannot be listed exhaustively in a dictionary (eg. World wide Web, Multiword Expressions). Whereas Light verb constructions are semi-productive, i.e. new light verb constructions might enter the language following some patterns (e.g. 'give a Skype call' on the basis of 'give a call').

Light Verb compounds are syntactically very flexible. They can manifest in various forms: the verb can be inflected, the noun can occur in its plural form and the noun can be modified. The nominal and the verbal component may not even be contiguous (eg. 'He gave me a very helpful advice').

**Methods of MWE identification**

1. Lowercase n-grams which occurred as links were collected from Wikipedia articles and the list was automatically filtered in order to delete non-English terms, named entities and non-nominal compounds etc.
2. Match: A noun compound is taken into consideration if it belongs to the list or it is composed of two or more noun compounds from the list.
3. POS rules: A noun compound candidate was marked if it occurred in the list and its POS-tag sequence matched one of the predefined patterns.
4. Suffix rule: The 'Suffix' method exploited the fact that many nominal components in light verb constructions are derived from verbs. Thus, in this case only constructions that contained nouns ending in certain derivational suffixes were allowed and for nominal compounds the last noun had to have this ending.
5. Most frequent method: This routine relied on the fact that the most common verbs function typically as light verbs (e.g. do, make, take, have etc.). Thus, the 15 most frequent verbs typical of light verb constructions were collected and constructions where the stem of the verbal component was among those of the most frequent ones were accepted.
6. Stem rule: In the case of light verb constructions, the nominal component is typically one that is derived from a verbal stem (make a decision) or coincides with a verb (have a walk).
7. Syntactic Information: Generally the syntactic relation between the verb and the nominal component in a light verb construction is verb-object.

### 3.1.2 Statistical Methods for Multiwords Extraction

A number of basic statistical methods can be used for extracting collocations from a given corpus [7] [19]. The corpus used for carrying out the experiments was a collection of The New York Times newswire for four months that consisted of 14 million words. Let us look at these methods and their corresponding applications for extracting multiwords.

### *3.1.2.3.    Frequency*

This is the simplest method for extracting collocations as it just retrieves the most frequent bigrams in the corpora. But this naive approach produced a lot of insignificant bigrams which are very frequent (*of-the*,*in-the* etc.) This difficulty can be easily overcome by applying a simple heuristic - pass the candidate phrases through a POS tagger and take only those combinations into considerations that have the probability of being phrases. The POStag structures that were taken into account were: AN, NN, AAN, ANN, NAN, NNN, NPN.

As we can see in Figure 3.1 even though it is a very simple method the results produced by this method was quite impressive.

| C($w^1 w^2$) | $W^1$ | $W^2$ | Tag Pattern |
|---|---|---|---|
| **11487** | New | York | AN |
| **7261** | United | States | AN |
| **5412** | Los | Angeles | NN |
| **3301** | Last | Year | AN |
| **3191** | Saudi | Arabia | NN |
| **2699** | Last | Week | AN |
| **2514** | Vice | President | AN |

**Figure 3.1**: Finding Collocations: Frequency Method [19]

### *3.1.2.4.* *Mean And Variance*

The above method for frequency works only for fixed phrases but there are words which stand in a flexible or variable length relationship length from one another. These are the words that appear with each other very frequently but can take any number of words in between.

Example: *knock...door*, this is a proper collocation even though there might be any number of words between *knock* and *door* depending on the structure of the sentence but *knock* is generally the verb associated with *door*.

In this method we calculate the mean and variance of the distance between two words. The variance is defined as:

$$s^2 = \frac{\sum_{i=1}^{n}(d_i - \bar{d})^2}{n-1}$$

Where 'n' is the number of times the two words co-occur, $d_i$ is the offset for co-occurrence 'i', and $\bar{d}$ is the sample mean of the offsets. If the offsets are same for most occurences the variance will be low and if the offsets differ highly for the occurences then the variance will be very high.

| s | d | Count | Word1 | Word2 |
|---|---|---|---|---|
| 0.43 | 0.97 | 11657 | New | York |
| 0.48 | 1.83 | 24 | Previous | Games |
| 0.15 | 2.98 | 46 | Minus | Points |
| 4.03 | 0.44 | 36 | Editorial | Atlanta |
| 4.03 | 0.00 | 78 | Ring | New |
| 3.96 | 0.19 | 119 | Point | Hundredth |
| 1.07 | 1.45 | 80 | Strong | Support |
| 1.13 | 2.57 | 7 | Powerful | Organizations |
| 1.01 | 2.00 | 112 | Rechard | Nixon |

**Figure 3.2**: Finding Collocations: Mean and Variance[19]

### *3.1.2.5.* *Hypothesis Testing*

The basic problem that we want to solve for collocation extraction is determining whether two words occur together more often than chance. Hypothesis testing is a classic approach in statistics for this type of problems. A null hypothesis $H_0$ is formed for this stating that the two words occur merely by chance. Now the probability of occurence of the two words given that $H_0$ is true is calculated, and then depending on this value of probability the null hypothesis is accepted or rejected.

### *3.1.2.5.1.* *The t-test*

The t-test looks at the mean and variance of a sample, where the null hypothesis is that the sample is drawn from a distribution with mean $\mu$. The test computes the difference between the observed and expected means, scaled by the variance of the data, and tells us how likely it is to get a sample of that mean and variance (or a more extreme mean and variance) assuming that the sample follows normal distribution.

$$t = \frac{\bar{x} - \mu}{\sqrt{s^2/n}}$$

Where $s^2$ is the sample variance, N is the sample size, $\mu$ is the mean of the distribution. If the t statistic is large enough we can reject the null hypothesis stating that the words are associated. For example, in the corpus, *new* occurs 15,828 times, *companies* 4,675 times, and there are 14,307,668 tokens overall.

*new companies* occurs 8 times among the 14,307,668 bigrams

$$H_0 : P(new companies) = P(new)P(companies)$$

$$= \frac{15828}{14307668} * \frac{4675}{14307668}$$

$$\approx 3.675 * 10^{-7}$$

The observed frequency of occurence of *new companies* is 8 in the corpus.

$$\bar{x} = \frac{8}{14307668}$$

Now applying the t-test:

$$t = \frac{\bar{x} - \mu}{\sqrt{s^2/n}}$$

$$\approx \frac{5.591*10^{-7} - 3.675*10^{-7}}{\sqrt{\frac{5.591*10^{-7}}{14307668}}}$$

$$\approx .999932$$

This t value of 0.999932 is not larger than 2.576, the critical value for $\alpha = 0.005$. So we cannot reject the null hypothesis that new and companies occur independently and do not form a collocation.

### 3.1.2.5.2. Hypothesis Testing of Differences

A variation of the basic t-test can be used to find words whose co-occurences best distinguish the subtle difference between two near synonyms. Figure 3.3 shows the words that occur significantly more often with *powerful* (the first ten words) and *strong* (the last ten words).
The formula of the basic t-test is modified as

$$t = \frac{\bar{x}_1 - \bar{x}_2}{\sqrt{s_1^2/n_1 + s_2^2/n_2}}$$

The application for this form of the t test is lexicography. Such data is useful to a lexicographer wanting to write precise dictionary entries that bring out the difference between *strong* and *powerful*.

| t | C(w) | C(strong w) | C(powerful w) | Word |
|---|---|---|---|---|
| 3.1622 | 933 | 0 | 10 | Computers |
| 2.8284 | 2337 | 0 | 8 | Computer |
| 2.4494 | 289 | 0 | 6 | Symbol |
| 7.0710 | 3685 | 50 | 0 | Support |
| 6.3257 | 3616 | 58 | 7 | enough |
| 4.6904 | 986 | 22 | 0 | Safety |

**Figure 3.3**: Hypothesis Testing Of Differences [19]

### 3.1.2.5.3. Pearson's Chi-Square Test

The t-test assumes that the probabilities of occurence are approximately normally distributed, which is not true in general. It is an alternative test that doesnot depend on the normality assumption. The essence of the test is to compare the observed frequencies with the frequencies expected for independence. If the difference between observed and expected frequencies is large, then we can reject the null hypothesis of independence.

| | $W_1$ = new | $W_1 \neq$ new |
|---|---|---|
| $W_2$ = companies | 8 (new companies) | 4667 (eg: old companies) |
| $W_2 \neq$ companies | 15820 (eg: new machines) | 14287181 (eg: old machines) |

**Figure 3.4:** Pearson's Chi-Square Test [19]

Figure 3.4 shows the observed frquency values for *new* and *companies*. On these values the test is applied. If the difference between observed and expected frequencies is large, then we can reject the null hypothesis of independence.

The $\chi^2$ statistic sums the differences between observed and expected frequencies, scaled by the magnitude of the expected values:

$$\chi^2 = \sum_{i,j} \frac{{O_{ij} - E_{ij}}^2}{E_{ij}}$$

Where i ranges over rows of the table, j ranges over columns, $O_{ij}$ is the observed value for cell and $E_{ij}$ is the expected value.

### 3.1.2.5.4. Likelihood Ratio

This test produces simply a number that tells us how much more likely one hypothesis is than the other. So it more interpretable than any other forms of hypothesis testing. Moreover, likelihood ratios are more appropriate for sparse data than the Chi-Square test.

For applying likelihood testing, let us consider the following two hypothesis:

$$Hypothesis1: P(w^2 | w^1) = p = P(w^2 | \neg w^1)$$

$$Hypothesis2: P(w^2 | w^1) = p_1 \neq p_2 = P(w^2 | \neg w^1)$$

Hypothesis1 is a formalization of independence whereas Hypothesis2 is a formalization of dependence. We calculate the log likelihood ratio as:

$$log_2(\lambda) = log_2 \frac{L(H_1)}{L(H_2)}$$

| $-2\log(\lambda)$ | $C(w^1)$ | $C(w^2)$ | $C(w^1w^2)$ | $W^1$ | $W^2$ |
|---|---|---|---|---|---|
| **-1291.42** | 12593 | 932 | 150 | Most | Powerful |
| **99.31** | 379 | 932 | 10 | Politically | Powerful |
| **82.96** | 932 | 934 | 10 | Powerful | Computers |
| **80.39** | 932 | 3424 | 13 | Powerful | Force |
| **57.27** | 932 | 291 | 6 | Powerful | Symbol |
| **51.66** | 932 | 40 | 4 | Powerful | Lobbies |
| **51.52** | 171 | 932 | 43 | Economically | Powerful |
| **51.05** | 932 | 43 | 4 | Powerful | Magnet |
| **50.83** | 4458 | 932 | 10 | Less | powerful |

**Figure 3.5**: Likelihood Ratio[19]

The Figure 3.5 shows the top bigrams consisting of *powerful* when ranked according to likelihood ratio.

### 3.1.2.5.5.  Relative Frequency Ratio

Ratios of Relative Frequencies between different corpora can be used to discover collocations that are characteristic of a corpus when compared to the other.

| Ratio | 1990 | 1989 | W[1] | W[2] |
|---|---|---|---|---|
| 0.0241 | 2 | 68 | Karim | Obeid |
| 0.0372 | 2 | 44 | East | Berliners |
| 0.0372 | 2 | 44 | Miss | Manners |
| 0.0399 | 2 | 41 | 17 | Earthquake |
| 0.0409 | 2 | 40 | HUD | officials |

**Figure 3.6**: Relative Frequency Ratio [19]

This approach is most useful for the discovery of subject-specific collocations. It can be used to compare a general text with a domain-specific text.

### 3.1.2.6.  Mutual Information

This is a method derived from information theory measures where we can find out how much information does the presence of one word gives about another word in the context. Informally, it is a measure of the company that a word keeps.

Mutual information (for two words, x and y) can be defined as:

$$I(x, y) = log_2 \frac{P(x'y')}{P(x')P(y')}$$

$$= log_2 \frac{P(x'|y')}{P(x')}$$

$$= log_2 \frac{P(y'|x')}{P(y')}$$

None of the statistical methods work very well for sparse data but Mutual Information works particularly badly in sparse environments because of the structure of the equation.

For perfect dependence (i.e. whenever they occur, they occur together):

$$I(x, y) = log_2 \frac{P(x'y')}{P(x')P(y')}$$

$$= log_2 \frac{P(x')}{P(x')P(y')}$$

$$= log_2 \frac{1}{P(y')}$$

The value of mutual information score gets inversely proportional to the frequency value of the bigram. So the bigrams that are rare in the corpus gets an artificially inflated mutual information score.

For perfect independence (i.e. their occurence together is completely by chance):

$$I(x, y) = log_2 \frac{P(x'y')}{P(x')P(y')}$$

$$= log_2 \frac{P(x')P(y')}{P(x')P(y')}$$

$$= log_2 1$$

$$= 0$$

It can be inferred that Mutual Information is a good measure of independence between two words but it is a bad measure for deciding the dependence between a bigram.

### 3.1.2.7. Comparative Analysis

We would like to present a comparative analysis in this section highlighting which method will be useful for what type of collocation.

- Frequency based method is simple and easy to implement hence it will be very useful for lightweight computations (Eg: Information Retrieval through search engines).
- Mean and Variance method can be used for terminological extraction and Natural Language Generation as it works well for variable length phrases.
- t-Test is most useful for ranking collocations and not so much for classifying whether a bigram is a collocation or not.
- Hypothesis Testing Of Differences is most useful for choosing between alternatives while generating text.

- Pearson's $\chi^2$ test is useful for identification of translation pairs among aligned corpora and measuring corpus similarity.
- Likelihood Ratios are more appropriate for sparse data than any other statistical method.

### 3.1.3 Word Association Measures

This is one of the very early attempts at collocation extraction by Kenneth Church and Pattrick Hanks (1990) [6]. They have generalized the idea of collocation to include **co-occurrence**. Two words are said to co-occur if they appear in the same documents very frequently.

For example: *doctor* and *nurse* or *doctor* and *hospital* are highly associated with each other as they occur together very frequently in a text.

The information theoretic measure, mutual information was used for measuring the word association norms from a corpus and then the collocations were produced.

#### *3.1.3.1.   Word Association And Psycholinguistics*

Word association norms are an important factor in psycholinguistic research. Informally speaking, a person responds quicker to a word hospital when he has encountered a highly associated word doctor before. In a psycholinguistic experiment a few thousand people were asked to write down a word that comes to their mind after each of the 200 words that were given to them. This was an empirical way of measuring word associations.

#### *3.1.3.2.   Information Theoretic Measure*

Mutual Information: If two words(x, y) have their probability of occurrence as P(x) and P(y) then their mutual information is defined as:

$$I(x, y) = log_2 \frac{P(x, y)}{P(x)P(y)}$$

Informally, mutual information compares the probability of x and y appearing together to, the probability of x and y occuring independent of each other. If there is some association between x and y then the mutual probability P(x,y) will be much greater than their independent probability P(x).P(y) and hence I(x,y)>>0. On the other hand, if there is no association between x and y then $P(x, y) \approx P(x).P(y)$, hence $I(x, y) \approx 0$.

The word probabilities P(x) and P(y) are estimated by counting the number of observations of x and y in a corpus (normalized by N, the size of the corpus).

Mutual probabilities, P(x,y) is estimated by counting the number of times x is followed by y in a window of w words, $f_w(x, y)$ (normalized by N, the size of the corpus). The window size allows us to look for different kinds of associations. Smaller window size identifies the fixed expressions whereas larger window size enables us to understand semantic concepts.

The association ratio is technically different from mutual information since in case of mutual information $f(x, y) = f(y, x)$ but that is not the case for association ratio because here linear precedence is taken into account.

### 3.1.3.3. Lexico-Syntactic Regularities

The association ratio is also useful to find out important lexico-syntactic relationships between verbs and their arguments or adjuncts. For example, consider the phrasal verb *set off*.

Using Sinclair's estimates

$$P(set) \approx 250*10^{-6}, P(off) \approx 556*10^{-6}$$
$$P(set, off) \approx 70/(7.3*10^6)$$

The mutual information for *set off* is:

$$I(set; off) = log_2 \frac{P(set, off)}{P(set)P(off)} \approx 6.1$$

From the above value we can infer that the association between *set* and *off* is quite large ($2^6$ i.e. 64 times larger than chance).

### 3.1.3.4. Importance Of Word Association

This was a pioneering approach towards extracting word associations. It extended the psycholinguistic notion of word association norm towards an information theoritic measure of mutual information. Informally, it helped us predict what word to look for if we have encountered some word. A lot can be predicted about a word by looking at the company that it keeps.

## 3.1.4 Retrieving Collocations From Text : XTRACT

Frank Smadja has implemented a set of statistical techniques and developed a lexicographic tool, Xtract to retrieve collocations from text [20]. As already stated, the definiton of collocation varies from one author to another.

According to the author, collocations have the following features:
- **Arbitrary** : They cannot be directly translated from one language to another as they are difficult to produce from a logical perspective.
- **Domain-dependent** : There are expressions which make sense only in a specific domain. These collocations will be unknown to someone not familiar with the domain.
- **Recurrent** : Collocations are not exceptional or chance co-occurences of words, rather they occur very frequently in a given context
- **Cohesive lexical clusters** : Encountering one word or one part of a collocation often suggests the probability of encountering the rest of the collocation as well.

The author has also classified collocations into three types:
- **Predicative Relations :** Two words are said to form a predicative relation if they occur very frequently in a similar syntactic structure (like, Adjective-Noun, Noun-Verb etc)
  For example : *make-decision* , *hostile-takeover*
- **Rigid Noun Phrases :** This involves uninterrupted, fixed sequences of words
  For example : *stock exchange,railway station*
- **Phrasal Templates :** Phrasal templates consist of idiomatic phrases consisting of one or more or no empty slots. These are generally used for language generation.
  For example : *Temperatures indicate yesterday's highest and lowest readings* is how generally a weather report begins.

### *3.1.4.1.   Xtract: The lexicographic tool for collocation extraction*

Xtract does a three stage analysis to locate interesting word associations in the context and make statistical observation to identify collocations. The three stages of analysis are:
- First Stage: statistical measures are used to retrieve from a corpus pair wise lexical relations.
- Second Stage: uses the output bigrams (of 1st stage) to produce collocations of n-grams.

- Third Stage: adds syntactic information to collocations retrieved at the first stage and filters out inappropriate ones.

The experiments were carried out on a 10million word corpus of stock market news reports.

### 3.1.4.1.1. Xtract: Stage One

Two words are said to co-occur if they are in a single sentence and if there are fewer than five words between them.

The words form a collocation if:
- They appear together significantly more often than expected by chance.
- Because of syntactic constraints they appear in a rigid way.

The algorithm used for extracting the bigrams forming collocations is:
1. Given a tagged corpus output all sentences containing a word w
2. Produce a list of words $w_i$ with frequency information on how w and $w_i$ co-occur.

$Freq_i$ (the frequency of appearance of $w_i$ with w in the corpus), POStag of $w_i$, $P^i_j (-5 \geq j \leq 5, j \neq 0)$ (frequency of occuring $w_i$ with w such that they are j words apart).

3. Analyze the statistical distribution and select interesting word pairs.

$$\text{Strength } (w, w_i) = k_i = \frac{freq_i - \bar{f}}{\sigma}$$

$\bar{f}$ and $\sigma$ are the average frequency and standard deviation of all the collocates of a word w

$$\text{Spread } (U_i) = \frac{\sum_{j=1}^{10}(p_i^j - \bar{p}_i)^2}{10}$$

If $U_i$ is small then the histogram will be flat implying that $w_i$ can be used at any position around w. Whereas if $U_i$ is large then the histogram will have sharp peaks implying that $w_i$ can be used only in some specific positions around w.

At the end of this stage a lexical relation corresponding to w is produced as output. It is of the form of a tuple ($w_i$,distance,strength,spread,j) verifying the following inequalities:

$$\text{Strength}= \frac{freq_i - \bar{f}}{\sigma} \geq k_0$$

$$\text{Spread} \geq U_0$$

$$p_j^i \geq \bar{p}_i + (k_1 * \sqrt{U_i})$$

Where $k_0, k_1, U_0$ are thresholds set manually.

### 3.1.4.1.2. Xtract: Stage Two

The second stage of Xtract produces collocations consisting of more than two words and also filters out some pairwise relations. The algorithm followed in stage two is given below.

1. Produce Concordances : Given a pair of words and the distance of the two words, produce all the sentences containing them in the specific position.
2. Compile and Sort : compute the frequency of appearance of each of the collocates of w
3. Analyze and Filter : a word or a POS is kept in the final n-gram at position if and only if

$$p(word[i] = w_0) \geq T$$

where T is a threshold set manually while performing the experiment

Some of the results after stage two are shown below:

| | | |
|---|---|---|
| Tuesday | the Dow Jones industrial average | rose 26.28 points to 2304.69 |
| | The Dow Jones industrial average | went up 11.36 points today. |
| …that sent | the Dow Jones industrial average | down sharply.. |
| Monday | the Dow Jones industrial average | was down 17.33 points to 2287.36… |
| …in | the Dow Jones industrial average | was the biggest since… |

**Figure 3.7** : Producing concordances for "the Dow Jones Industrial Average"[20]

| | |
|---|---|
| The NYSE composite index of all its listed common stocks | fell 1.76 to 164.13 |
| The NYSE composite index of all its listed common stocks | fell 0.98 to 164.97 |
| The NYSE composite index of all its listed common stocks | fell 0.91 to 164.98 |
| The NYSE composite index of all its listed common stocks | rose 0.76 |
| The NYSE composite index of all its listed common stocks | fell 0.33 to 170.63 |

**Figure 3.8**: Producing the "NYSE's composite index of all its listed common stocks " [20]

In stage two of Xtract:
- Phrasal templates are also produced in addition to rigid noun phrases
- Produces the biggest possible n-gram
- Relatively simpler way of producing n-grams

### 3.1.4.1.3. Xtract: Stage Three

In stage three of Xtract the collocations produced in stage one are analyzed and the syntactic relationship between them is established otherwise they are rejected.

1. Produce Concordances : Given a pair of words and the distance of the two words, produce all the sentences containing them in the specific position.
2. Parse : For each sentence produce set of syntactic labels
3. Label and Filter : count the frequencies of each possible label identified for the bigram (w,wi) and accept if and only if

$$p(label[i] = t) \geq T$$

Where T is a threshold defined manually while performing the experiment

For example: If after the first two stages of Xtract the collocation *make-decision* is produced then in the third stage it is identified as a *verb-object* collocation. If no such relationship can be established then such collocations are rejected.

### 3.1.4.2. Analysis Of Xtract

The precision and recall value of Xtract are 80% and 94% respectively. An observation that can be made from the results of Xtract is that the extracted collocations are domain dependent. Hence the domain and size of the corpus has heavy influence on the type of collocations extracted from it. This work showed a nice method of extracting 'n-grams' and by adding syntax to the collocations it could explain the syntactic relationships between the colloactes as well.

### 3.1.5 Collcation Extraction By Conceptual Similarity

This is a method suggetsed in [14] where the author uses Wordnet to find out the conceptual similarity between different words. It is observed that in spite of the similarity between words due to the arbitrary nature of collocations only one of the many possible synonyms of a word a candidate phrase prefers one word over another. From this point of view collocation can be redefined as:

A pair of words is considered to be a collocation if one of the words significantly prefers a particular lexical realization of the concept the other represents. Consider the following examples:

| Correct Expression | Incorrect Expression |
|---|---|
| many thanks | several thanks |
| emotional baggage | emotional luggage |
| strong coffee | powerful coffee |
| tap water | pipe water |

Table 3.1: Collocation Preference

For example, *coffee* significantly prefers *strong* over *powerful* and similarly the other examples. In this new outlook there's an inherent directionality as each candidate phrase prefers one synonym over another. So this is termed as **collocation preference**.

The authors studied the usages of two similar words, *baggage* and *luggage:*
1. 2 million parsed sentences of BNC were searched for occurrences of the synonyms *baggage* and *luggage*. If the difference of their occurrence for a particular word was greater than 2 then that bigram was taken into account.
2. For each such bigram obtained in step1, Alta Vista search was used to find occurrences of it in the world wide web.
3. Details of collocation according to CIDE(Cambridge International Dictionary Of English) was used as standard of judgment.

Figure 3.9 shows the difference in usage for the two synonyms *baggage* and *luggage*.

| Word | BNC | Alta Vista | CIDE | Collocation |
|---|---|---|---|---|
| **allowance** | B 5 0 | B 3279 502 | B | baggage allowance |
| **area** | B 3 1 | B 1814 1434 | | ? baggage area ? |
| **car** | B 4 0 | B 3324 357 | B | baggage car |
| **compartment** | L 1 3 | L 2890 5144 | L | luggage compartment |
| **label** | L 0 6 | L 103 333 | L | luggage label |
| **rack** | L 0 8 | L 164 14773 | L | luggage rack |

**Figure 3.9**: Collocational Information for 'baggage' and 'luggage'[14]

### *3.1.5.1.* *Collocation Graph*

Collocation graphs are diagrammatic representation of the different senses represented by a word and the arcs are used to denote colloactional preferences described as follows.

### *3.1.5.1.1.* *Concept Set*

A collocation graph consists of two or more **concept nodes** that represent the senses that a word has according to the Wordnet. For a word *w* the concept set *C(w)* is defined as:

$$C(w) = \{S_i : w \in S_i\}$$

For example, the word *information* has five meanings according to the Wordnet. So its concept node will have five entries, one for each of the meanings.

### *3.1.5.1.2.* *Intersection Of Concept Sets*

If two words are synonyms in some sense i.e. they share a sense in common then their concept nodes will have an intersection and that sense (common to both of them will be present in the intersection).

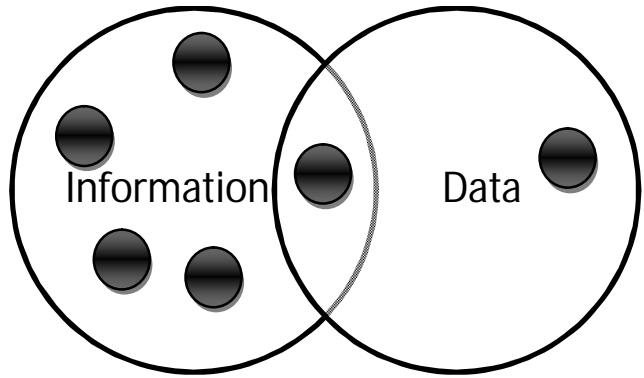

**Figure 3.10:** Intersection Of Concept Sets for information and data

Figure 3.10 shows the intersection of the concept sets of *information* and *data*.

### *3.1.5.1.3.    Collocation Preference*

Concept nodes in a concept graph are connected by collocation arcs to show the preference that is being exhibited due to the property of collocations. The direction of the arc represents which word is expressing preference for which word.

### *3.1.5.1.4.    Intersection Graphs*

While trying to determine significant collocations the concept nodes for the synonyms are drawn. They have one or more senses in common. A candidate phrase is said to exhibit collocational preference if it is expressing more preference for one word than the other for representing the same sense. This is denoted by a directed preference arc in the collocation graph and the arc passes through the preffered word first. This is shown as an example for *emotional baggage* and *emotional luggage* in **Figure 3.11**.

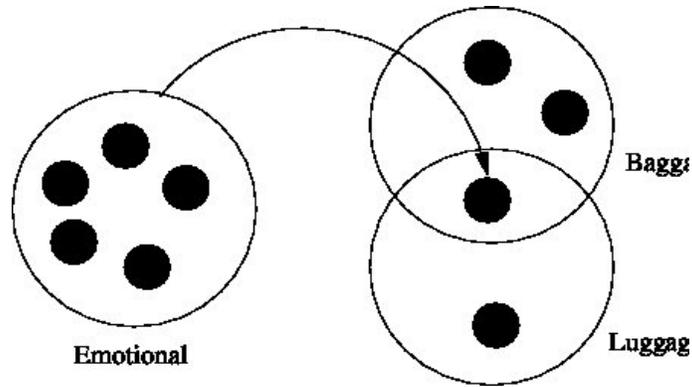
**Figure 3.11**: Collocational Preference

### *3.1.5.2.   Poly-Collocations*

It might be possible for a word to express preference for another word in more than one of its synsets. These are termed as **poly-collocations**. Depending on whether sense information is available or not a variety of configurations are possible for the collocation graph.

### *3.1.5.3.   Anti-collocations*

A synonym set with respect to a particular target phrase can be classified into three disjoint sets :

• The words which are frequently used with the target word (**Collocations**).

• The words which are generally not used with the target word but do not lead to unnatural reading.

• The words which are never used with the target word (**Anti Collocation**).

The knowledge of anti-collocations will be very much helpful for natural language generation and foreign language learners.

Example: *strong drugs, *powerful coffee

### *3.1.5.4.   Formalization*

The algorithm takes a sequence of bigrams $p^1, p^2 ... p^N$ as input.

- The occurence count for each such pair is defined as :

$$C(a,b) = \sum_{i=1}^{n}(\delta(p^i = \prec a,b \succ))$$

where, $\delta(x) = 1$, if x is true and is 0, if x is false.

- The co-occurence set of a word w is defined as:

$$cs(w) = \{v : c(w,v) > 0\}$$

- Wordnet is defined as a set of synsets, W. Candidate collocation synset of a word w is defined as:

$$CCS(w) = \{S \in W : |S \cap cs(w)| > 2\}$$

So each candidate collocation synset S,(for a word w) consists of atleast two elements whose co-occurence count is non-zero.

- Most frequently co-occurring element of a synset and its frequency are defined as:

$$w' = argmax\, c(w,v) \quad f' = argmax\, c(w,v)$$

- Collocation strength is defined as $f' - f''$ where $f''$ is the second highest frequency in the synset.

### 3.1.5.5. Analysis

The idea presented in the paper looks promising and since the work is at the semantic level it is more intuitive and easy to connect to how a human mind works in reality.

The future work needs to focus on improving the basic algorithm in particular aspects :
  • The idea of synonym set can be extended to concept set.
  • Experiments need to be conducted for synsets other than Nouns.
  • Morphological processing need to be done.
  • Some thesaurus can be used along with Wordnet.

## 3.1.6 Verb Phrase Idiomatic Expressions

An idiom can be defined as a speech form or an expression of a given language that is peculiar to itself grammatically or cannot be understood from the individual meanings of its elements.

For example: by and large, spill the beans, shoot the breeze, break the ice etc.

These are very typical to a language and evolve over time. Even within a language they vary from one dialect to another.

Idioms don't follow some general conventions among its class. Like,some of them might allow some form of verbal inflection (*shot the breeze*) whereas some might be completely fixed (*now* *and then*). The idioms that are perfectly grammatical are difficult to be identified as an idiom having idiosyncratic meaning as opposed to its similar structures (*shoot the breeze* and *shoot the bird*).

The authors have looked into two closely related problems confronting the appropriate treatment of Verb-Noun Idiomatic Combinations(where the noun is the direct object of the verb) [8]:

- The problem of determining their degree of flexibility
- The problem of determining their level of idiomaticity

### 3.1.6.1. Recognizing VNICs

Even though VNICs vary in their degree of flexibility on the whole, they contrast with compositional phrases (which are more lexically productive and appear in a wider range of syntactic forms). Hence the degree of lexical and syntactic flexibility of a given verb+noun combination can be used to determine the level of idiomaticity of the expression. The authors have tried to measure the lexical and syntactic fixedness of an expression by a statistical approach to determine whther it is an idiom or not.

### 3.1.6.2. Analysis

Idioms form a very interesting part of natural language but due to its pecularity and arbitrary nature it has been side-stepped by the NLP researchers for long. The authors have tried to provide an effective mechanism for the treatment of a broadly documented and crosslinguistically frequent class of idioms, i.e., VNICs. They have done a deep examination of several linguistic properties of VNICs that distinguish them from similar literal expressions. Novel techniques for translating such characteristics into measures that predict the idiomaticity level of verb+noun combinations have also been proposed.

## 3.2. Study of an Ongoing Project: MWEToolkit

Multiword Expression Toolkit (mwetoolkit) is developed for type and language-independent MWE identification [4]. It is a hybrid system for detecting multiwords from a corpus using rule based as well statistical association measures. The toolkit is a open source software can be downloaded from [sf.net/projects/](sf.net/projects/).

## 3.2.1 MWEToolkit System Architecture

Given a text corpora the toolkit filters out the MWE candidates from the corpora. The different phases present in the toolkit to achieve this goal are:

1. Preprocessing the corpus: Preprocess the corpus for lowercase conversion, lemmatization and POS tagging (using Tree tagger).
2. Extract 'ngrams' depending on the predefined POS patterns.
3. For each of these bigrams take into account their corpus count as well as the web count (number of pages in which the particular bigram is present) using Google and Yahoo
4. Apply some Association Measures (statistical) to filter out the candidates.
    i. The corpus containing the N word tokens is indexed and from that index the counts of the tokens are estimated. Using the index, individual word counts, $c(w_1)$, $c(w_2)$……$c(w_n)$ and the overall ngram count $c(w_1w_2…w_n)$ is computed.
    ii. The expected N gram is computed if words occurred just by chance
    $$E \approx \frac{c(w_1).c(w_2).c(w_3)………c(w_n)}{N^{n-1}}$$
    iii. Using the above information four Association Measures are computed
    - Maximum Likelihood Estimator
    $$mle = \frac{c(w_1w_2………w_n)}{N}$$
    - Dice's coefficient
    $$dice = \frac{n * c(w_1w_2………w_n)}{\sum_{i=1}^{n} c(w_i)}$$
    - Pointwise Mutual Information
    $$pmi = \log_2 \frac{c(w_1w_2………w_n)}{E(w_1w_2………w_n)}$$
    - Students' t-score
    $$t-score = \frac{c(w_1w_2………w_n) - E(w_1w_2………w_n)}{\sqrt{c(w_1w_2………w_n)}}$$
5. Once each candidate has a set of associated features, an existing machine learning model can be applied to distinguish true and false positives or a new model can be designed by assigning a class to the new candidate set.

## 3.2.2 Using Web as corpora

Another novel aspect of the system is, it uses web count of MWEs as a feature for their Machine Learning model. Let us look a bit more closely and analyze the advantages and disadvantages of using web as a corpus.

**Issues:**

- Web counts are "estimated" or "approximated" as page counts, whereas standard corpus counts are the exact number of occurrences of the n-gram.

- In the web count, the occurrences of an n-gram are not precisely calculated in relation to the occurrences of the $(n-1)$-grams composing it.
  For instance, the n-gram "the man" may appear in 200,000 pages, while the words "the " and "man" appear in respectively 1,000,000 and 200,000 pages, implying that the word "man" occurs with no other word than "the".

- Unlike the size of a standard corpus, which can be easily computed, it is very difficult to estimate how many pages exist on the web and especially because this number is always increasing.

**Advantage:**

- In spite of the issues, the biggest advantage of the web is its **availability**, even for resource-poor languages and domains. It is a free, expanding and easily accessible resource that is representative of language use, in the sense that it contains a great variability of writing styles, text genres, language levels and knowledge domains.

- The web can minimize the problem of sparse data. Most of the statistical methods suffer due to the sparsely distributed data in the corpus. Web can lend a hand for dealing with this problem. Due to the sheer volume of data present on the web, it can assist us to distinguish rare occurrences from invalid cases.

# Chapter 4

# Multiword Extraction Engine

In this section we are going to describe the Multiword Extraction Engine that has been developed in IIT Bombay to extract multiword expressions from English as well as all major Indian languages. This is a hybrid system employing both linguistic rules and statistical methods for MWE extraction.

## 4.1. System Overview

The overview of the MWE extraction pipeline is shown in Figure 4.1. POS tagged corpus is fed to the system.

- Regular Expression (RegEx) filter, filters out the MWE candidates on the basis of pre-specified regular expression patterns.
- Partial Reduplication filter is applied to detect occurrences of both meaningful and non-meaningful partial reduplication.
- The filtered out candidates are passed to the Linguistic Filter, which itself is composed of three filters namely Vector Verb & Verbalizer Filter, Named Entity Filter and Hyphenation Filter.
    - In case the candidate belongs to Verb + Verb category or Noun + Verb category, candidate is passed to the very first filter of the Linguistic Filter module.
    - To filter out the noise created by the Named Entity, candidates are then passed to the Named Entity Filter.
    - The Hyphenation filter filters in the candidates which have hyphen ("-") in between them, as they are most likely to be MWE.
- Complex Predicate Filter is applied to computationally detect conjunct verbs using some heuristics.
- After the Linguistic Filter, the candidates are checked for any semantic relationship between themselves. At this step it is checked if the constituents of a bigram are synonyms or antonyms of each other or they belong to the same class of concepts using Wordnet.

- The candidates are then ranked using the Statistical measures including Point-wise Mutual Information (PMI), Dice coefficient and Log-Likelihood Algorithms. At the end of this step, a combined ranked list is generated by the engine.
- Finally the ranked list of the candidates is analyzed by a lexicographer. This is the manual filtering step, which provides the user options to browse the list and analyze the candidates to ascertain whether they are true MWEs or not. They also have the option to detect false positives and false negatives. After this step a Gold Standard MWE List is produced.

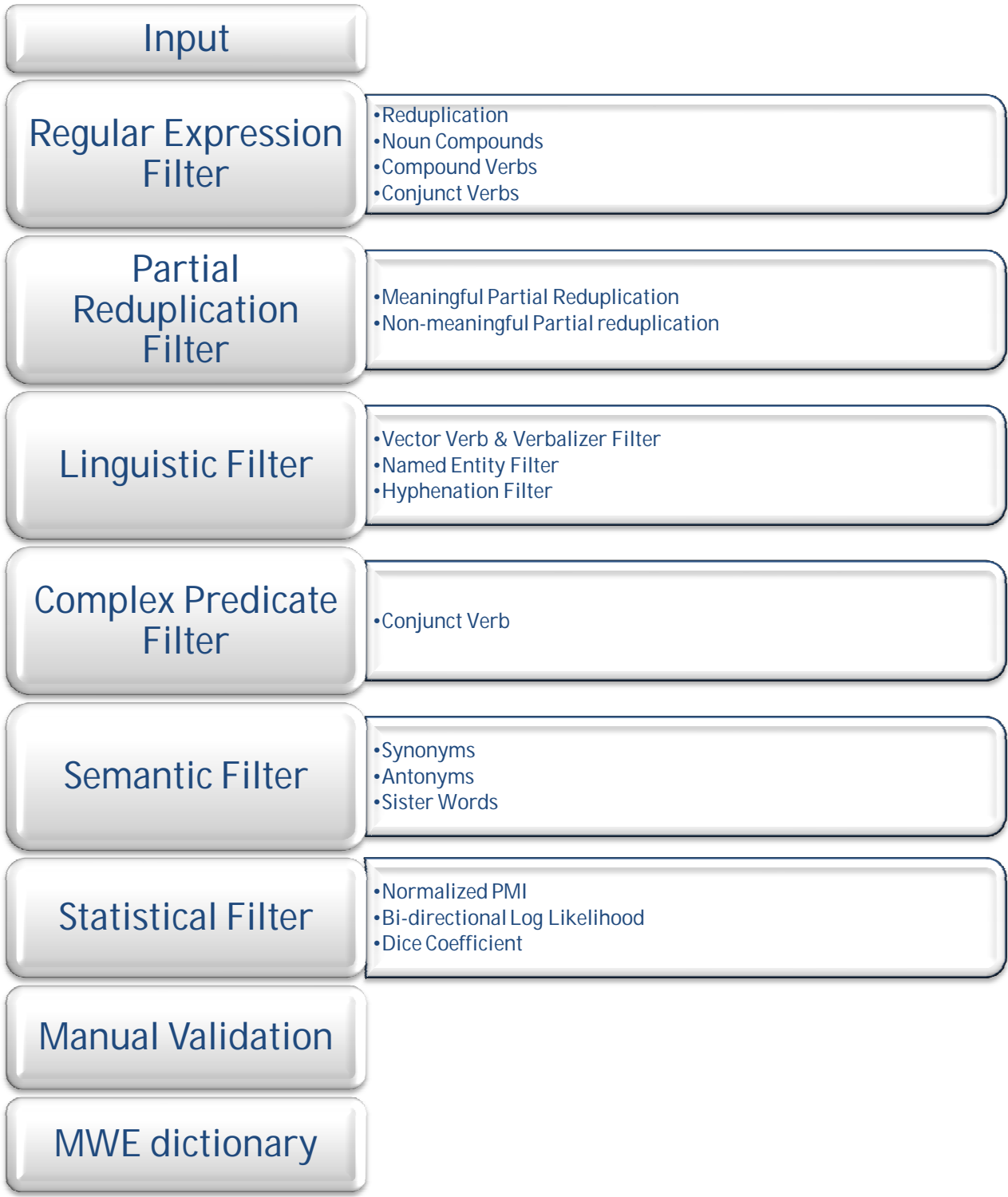

**Figure 4.1**: MWE Engine Pipeline

### 4.1.1 System Features

1. This is a completely multilingual system that requires only raw text as input and can extract the multiword expressions from the text.
2. It is highly scalable for large size of data without much memory overhead because it uses Lucene's index as data structure.
3. A multilingual web-based service has been developed which will be hosted at IIT Bombay in order to facilitate the use of this engine by all research groups across India.

### 4.1.2 Integrated Resources

- **Linguistic Resources**: Lists of Vector verbs, Verbalizer and Named Entities are required for each language to apply the linguistic filter.
- **Wordnet**: The system is integrated with Princeton Wordnet (English) and Indo Wordnet (Indian Languages) to detect semantic relationship between the constituents of a candidate.

## 4.2. MWE Extraction Engine Pipeline

In this section different stages of the pipeline are explained in detail. The required input for each stage, the output produced at the end of that stage and their functionalities are analyzed.

### 4.2.1 Regular Expression Filter

This is the first filter in MWE extraction pipeline. It takes the POS tagged data as input and searches for some predefined patterns. Since multiword expressions are very heterogeneous in nature the objective of this step is to generate as many candidates as possible. Since this is the first filter of the engine the goal here is to reduce the false negatives as much as possible. We take bigrams to pentagrams into account and have defined patterns for them separately.

**For Bigrams:**

#### *4.2.1.1.     Reduplication*

Here we check for repetition of a word irrespective of the tags because most of the POS taggers fail to tag reduplications properly.

**Pattern: word1_POS1 word2_POS2**                      where, word1=word2

Example:

                               Knock_VB knock_VB

                    Gloss: The sound of knocking at door

### *4.2.1.2.    Compound Noun*

For compound nouns we take all types of Noun-Noun compounds into account.

**Pattern: word1_Noun word2_Noun**                where                Noun= **NN|NNP|NNC|NNPC**

Example:

                               railway_NN station_NN

                               Gloss: railway station

                               शिव_NN मंदिर _NN (Hindi)

                               Transliteration: Shiva mandir

                               Gloss: Shiva temple

                               Translation: Temple of Shiva

### *4.2.1.3.    Compound Verb*

Compound verbs are formed when two verbs appear consecutively in a sentence.

**Pattern: word1_Verb word2_Verb**       where Verb= **VB|VBD|VBG|VBP|VBN|VBZ|VM**

Example:

                               चला_VB गया_VM (Hindi)

                               Transliteration: chala gaya

                               Gloss: has gone

### *4.2.1.4.    Conjunct Verb*

Syntactically conjunct verbs are Noun+Verb combinations.

**Pattern: word1_Noun word2_Verb**

Example:

                               सलाह_NN देना_VM

Transliteration: salaah dena

Gloss: advice give

Translation: to give advice

**For Trigrams, Quadra grams, Pentagrams:**

### *4.2.1.5.  Noun Compounds*

Noun compounds can merge together and form a new larger noun compound. Hence all noun-noun combinations were taken into account.

Example:

Science fiction writer

### *4.2.1.6.  Adjective + Noun Compounds*

It has been observed that a noun compound preceded by an adjective is very frequently a terminology or a collocation

Example:

Red blood corpuscle

## 4.2.2 Linguistic Filter

Linguistic Filter exploits linguistic knowledge to filter out the MWE candidates. These filters are created on the basis of observations on the languages and MWE Analysis hence it is language dependent. The filer contains three sub-filters.

### *4.2.2.1.  Vector Verb + Verbalizer Verb Filter*

Linguistic composition of Compound Verb and Conjunct Verb is

**Compound Verb** = Verb1 + Verb2 = Polar Verb + Vector Verb

- **हस उठना** :(Polar Verb) + (Vector Verb)

    **Transliteration**: has uthna

    **Translation:** laugh out

    **Gloss**: laugh up

**Conjunct Verb** = Noun + Verb = Noun + Verbalizer Verb

- **सलाह देना** :(Noun) + (Verbalizer)

    **Transliteration**: salah dena
    **Translation**: advice
    **Gloss**: advice give

Both of these, Vector verb and Verbalizer verb, are limited in number for Indian Languages. For a Compound Verb and Conjunct Verb candidate to be a MWE, its vector verb or verbalizer verb has to be from the list of vector verb or verbalizer verb list defined in the Language. Given a list of such verbs for a language verb compounds containing verbs from that list only are considered for further evaluation. The purpose of this filter is to reject the false positive compound verb and conjunct verb Bigram candidates

### *4.2.2.2.    Named Entity Filter*

Named Entities are very important part of the MWEs. As we go from Bi-gram to Pent-gram in Indian languages, the probability of MWE candidate being a Named Entity increases and most of the Quad-gram and Pent-gram candidates being extracted are Named Entities.

*Example*:

Indira Gandhi International Airport

In Bi-grams and Tri-grams these Named Entity candidates are causing lots of false positive candidates to occur. As Named Entity candidates are Noun + Noun combination, which combines with the other noun terms, coming just after them or just before them, creates false positive candidate for MWE.

Named Entity filters filter out the candidates which has a part of a Named Entity in them. A lower score is given to these words while ranking. The purpose of Named Entity filter is to filter out false positive candidates caused by Named Entities and give them low weight-age while ranking.

### *4.2.2.3.    Hyphenation Filter*

In Indian Languages when hyphen, "-", occurs in-between words, most of the times they are Multi-word Expressions.

*Example*:

- चतुर-चालाक (Hindi)

    **Transliteration:** Chatur-Chalak
    **Gloss:** smart-clever
    **Translation:** smart

These words are considered as single word by the RegEx filter, but are mainly MWE's, which when combine with the other words create false positive candidates.

The motivation of this filter is to filter in the words with hyphen in them, so as to decrease false negatives, as most of the words with hyphen are the MWEs. A higher weight-age is given to those words while ranking.

All these three filters combine to form the Linguistic filter, which decreases the number of false positive MWE candidates and provides the MWE candidates which have high probability of being a MWE.

### 4.2.3 Complex Predicate Filter

Complex predicate is a noun, a verb, an adjective or an adverb followed by a light verb that behaves as a single unit of verb. Complex predicates (CPs) are abundantly used in Hindi and other languages of Indo Aryan family. Detecting and interpreting CPs constitute an important and a somewhat difficult task. Most of the work that has been presented in the literature for detecting CPs are language dependent and uses a list of light verbs for the specific language.

We use a completely automated, language independent methodology, based on a lexical resource (IndoWordnet) for detecting conjunct verbs. Complex Predicates are abundant in Indo Aryan languages and their characteristics are also similar. The method described in this section is generic and is suitable for all such languages.

A conjunct verb is a multi-word expression (MWE) where a noun is followed by a light verb (LV) and the MWE behaves as a single unit of verb. The CP in a sentence syntactically acts as a single lexical unit of verb that has a meaning distinct from that of the LV i.e. N+V combination behaves as a 'single semantic unit' where the verb loses its individual meaning. For eg:

- तसल्ली देना (Hindi)

  **Transliteration:** tasalli dena
  **Gloss:** assurance give
  **Translation:** assure

- निष्कर्ष निकलना (Hindi)

  **Transliteration:** nishkarsh nikalna
  **Gloss:** summary extract
  **Translation:** summarize

- प्यार होना (Hindi)

  **Transliteration:** pyar hona
  **Gloss:** love happen
  **Translation:** love

- गर्व होना (Hindi)

  **Transliteration:** garv hona
  **Gloss:** pride happen
  **Translation:** proud

We use Wordnet and ontology to study the characteristics of the nouns and verbs to formulate when does such a combination makes complex predicate. In a conjunct verb, the meaning of the light verb is lost i.e. in '*तसल्ली देना*' none of the senses of the verb '*देना*' is used. According to ontology, there can be three types of verbs:

  i. Verb of Action (VOA)
  ii. Verb of State (VOS)
  iii. Verb of Occur (VOO)

'*देना*' is a Verb of Action. VOA generally requires an 'object' whereas 'तसल्ली' is an abstract noun and not an object. Hence the 'selectional preference' of the verb is not met and it is not used in its usual sense, therefore forming a conjunct. We look into the ontological categories of Nouns and Verbs and check the combinations which form conjuncts. Our findings are:

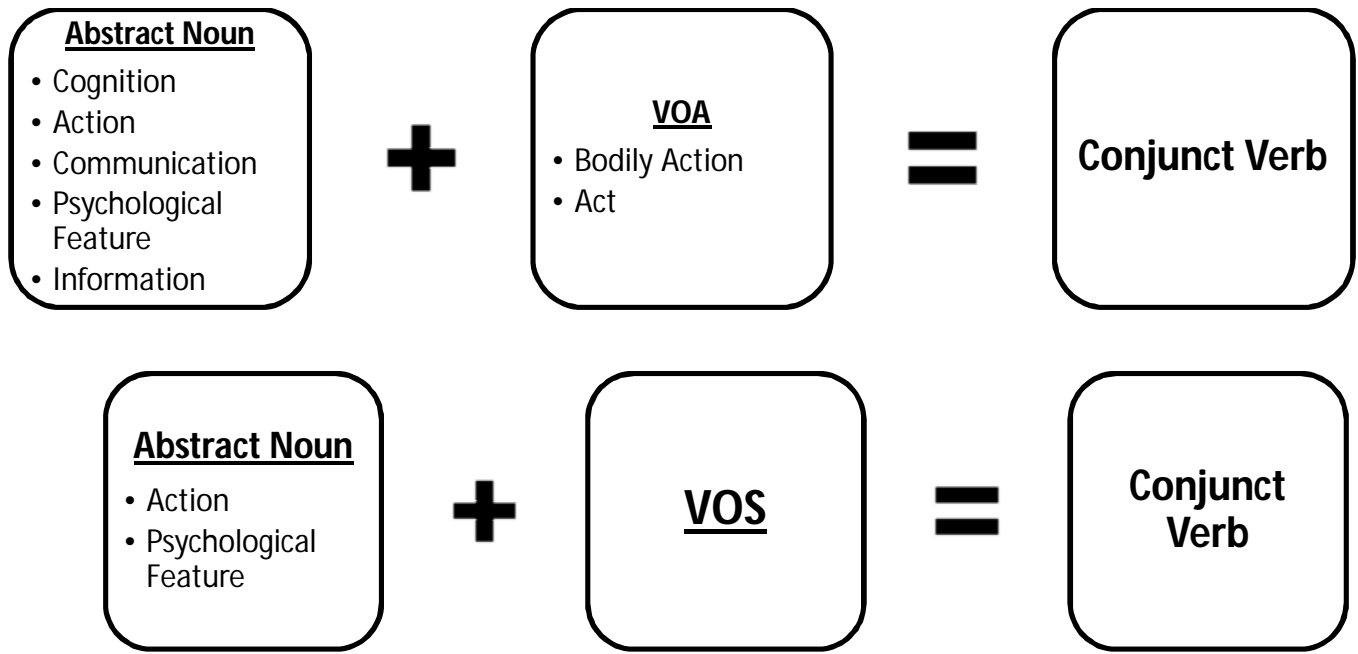

Table 4.1 shows some of the candidates for Conjunct Verb extracted by our system on a Hindi corpus.

| जोश_NN आना_VM (josh aana) | सहमति_NN दे_VM (sahamati de) |
| --- | --- |
| आस्था_NN उठ_VM (astha uth) | रोशनी_NN फूट_VM (roshni foot) |
| परिवर्तन_NN आना_VM (parivartan aana) | दर्जा_NN मिलना_VM (darja milna) |
| पसंद_NN आना_VM (pasand aana) | वरदान_NN मिला_VM (vardan mila) |
| सवाल_NN उठ_VM (sawaal uth) | नियंत्रण_NN रखना_VM (niyantran rakhna) |
| कदम_NN उठा_VM (kadam uth) | प्रस्ताव_NN रखना_VM (prastav rakhna) |
| बोझ_NN उठाना_VM (bojh uthana) | विचार_NN रखना_VM (vichar rakhna) |
| चेहरा_NN उतर_VM (chehra utar) | बाजी_NN लगाना_VM (baazi lagana) |
| उदाहरण_NN छोड़_VM (udaharan chhod) | हिसाब_NN लगाना_VM (hisaab lagana) |
| सलाह_NN दे_VM (salaah de) | सहारा_NN देना_VM (sahaara dena) |

**Table 4.1**: Observation Results for Conjunct Verbs

## 4.2.4 Semantic Filter

This module is used to check for semantic relationship between the constituents of a bigram. Wordnet is used as the primary resource for this.

For a bigram we search the Wordnet of that language to check for all synsets which contain the words. After retrieving the synsets, we check if there exists any synonymous, antonymous relationship between them. To check whether the words belong to the same class or not we check if they are sister words i.e. they share a common direct hypernym.

### *4.2.4.1.    Synonym*

- रिश्ते नाते (Hindi)

    **Transliteration:** rishte naate
    **Gloss:** relationship connection
    **Translation:** relationships

### *4.2.4.2.    Antonym*

- जीना मरना (Hindi)

    **Transliteration:** jeena marna
    **Gloss:** living dead
    **Translation:** everything

- আকাশ পাতাল (Bengali)

    **Transliteration:** akash patal
    **Gloss:** sky underground
    **Translation:** a lot

### *4.2.4.3.    Sister Words*

- भाई बहेन (Hindi)

    **Transliteration:** bhai bahen
    **Gloss:** brother sister
    **Translation:** everyone

## 4.2.5 Partial Reduplication Filter

This is the case where a word is partially duplicated. It can further be classified in two categories.

- ***Meaningful Partial Reduplication:*** Both the components are meaningful and form a rhyming pattern.

    - जाना माना (Hindi)

        **Transliteration:** jaana maana
        **Translation:** known acknowledged
        **Gloss:** renowned

    - चलते फिरते (Hindi)

        **Transliteration:** chalte firte
        **Translation:** while walking

We use Wordnet for detecting such patterns. We check if both the words have a matching suffix (implying that they form a rhyming pair) and both the words are meaningful, hence exist in a lexical database (like, wordnet).

Ideally, words forming a meaningful partial reduplication should have the same POS tag but it is difficult for a POS tagger to tag such expressions correctly. When we incorporated the restriction on POS tags, it reduced the recall due to errors from the POS tagger. Constituent words should also have the same number of phonemes in order to make a rhyming pattern. A point to note here is that, for Indian languages the number of phonemes and the number of characters in the word are not the same. Since this is a generic system, if we want to match the number of phonemes, we'll have to keep specific modules for each language.

Another challenge is, most of the Indian languages have rich morphology and inflected forms are not present in Wordnet. In order to search an inflected word in wordnet, we need to do morphological analysis. To the best of our knowledge morphological analyzer does not exist for all Indian languages at present. We solved this problem by developing a generic lemmatizer using wordnet, the details of which are described in Chapter 8. We use this lemmatizer to identify the root form of the word and determine whether it is meaningful or not. For eg:

- Input Combination: चलते फिरते (chalte firte)

- Stems: चल फिर (chal fir)

- Lemmatized Forms: चलना फिरना (chalna firna)

We first take the input and each word is given to the lemmatizer. The lemmatizer finds the root by maximal string matching with words from wordnet stored in a trie. Once we get the root forms we extract all the lemmas formed from that root and check if any of these lemmas for 1st word and 2nd word share the same suffix.

- ***Non-meaningful Partial Reduplication:*** In non-meaningful partial reduplication the first word is meaningful whereas the second word is a rhyming variation of the first word. For the construction of the second word generally three cases are possible
  i. Change of the first vowel or the matra attached with first consonant
  ii. Change of consonant itself in first position
  iii. Change of both matra and consonant

If the two words differ by only the first character and the second word is not present in a lexical database (we use Wordnet) then we consider it to be a case of Non-meaningful partial reduplication.

- चाय वाय (Hindi)

  Transliteration: chaye vaye
  Translation: tea

- मिलता जुलता (Hindi)

  Transliteration: milta julta
  Translation: similar

## 4.2.6 Statistical Filter

We employ statistical methods for detecting collocations. Collocations are the type of multiword expressions which are statistically idiosyncratic for a language and are not governed by any linguistic rule. We use three statistical methods for detecting collocations, namely,

- Normalized Point-wise Mutual Information
- Bidirectional Log-Likelihood Ratio

- Dice Coefficient

### 4.2.6.1. Normalized Point-wise Mutual Information

Point-wise Mutual Information is being used to statistically rank the MWE candidates on the basis of shared information score being calculated using the PMI formulae as mentioned below:

$$I(w_1, w_2) = log_2 \frac{p(w_1 w_2)}{p(w_1) * p(w_2)}$$

Where, $I(w_1,w_2)$ is the amount mutual information shared between words $w_1$ and $w_2$,

$p(w_1 w_2)$ is the probability of $w_1$ and $w_2$ occurring together,

$p(w_1)*p(w_2)$ is the probability of the words occurring in the corpus independently.

PMI is roughly a measure of how much one word tells us about the other and gives an idea of the 'dependence' between the words. If we look more closely at the formula, we'll see that when the two words are completely dependent on each other i.e. $w_1$ occurs only when it is followed by $w_2$, the PMI formula reduces to,

$$I(w_1, w_2) = log_2 \frac{p(w_1 w_2)}{p(w_1) * p(w_2)} = log_2 \frac{p(w_1)}{p(w_1) * p(w_2)} = log_2 \frac{1}{p(w_2)}$$

As the word combination gets rarer its PMI score gets higher. Bigrams composed of low-frequency words will receive a higher score than bigram composed of high-frequency words but it is not necessarily true that low frequency occurrences are of higher importance.

We normalize the original Point-wise Mutual Information formula to give importance to the frequency of occurrence. We also extend the PMI formula, which is for Bi-grams, to apply it to Tri-gram, Quad-gram and Pent-gram.

### 4.2.6.1.1. Bi-gram NPMI

$$I_{bi} = log_2 \frac{p(w_1 w_2)^2}{p(w_1) * p(w_2)}$$

Where, $I(w_1,w_2)$ is the amount mutual information shared between words $w_1$ and $w_2$,

$p(w_1 w_2)$ is the probability of $w_1$ and $w_2$ occurring together,

$p(w_1)*p(w_2)$ is the probability of the words occurring in the corpus independently.

We expand the formulas of (n-1) gram to find out PMI score for n-gram.

### *4.2.6.1.2.  Tri-gram NPMI*

$$I_{tri} = log_2 \frac{p(w_1w_2w_3)^2}{p(w_1w_2) * p(w_2w_3)}$$

Where, $I_{tri}$ is the amount mutual information shared between $w_1$, $w_2$ and $w_3$,

$p(w_1w_2\,w_3)$ is the probability of $w_1$, $w_2$ and $w_3$ occurring together,

$p(w_1w_2)\,*p(w_2w_3)$ is the probability of occurrence of the bigram $w_1w_2$ independent of the probability of the occurrence of the bigram $w_2w_3$ in the corpus.

### *4.2.6.1.3.  Quad-gram NPMI*

$$I_{quad} = log_2 \frac{p(w_1w_2w_3w_4)^2}{p(w_1w_2w_3) * p(w_2w_3w_4)}$$

Where, $I_{quad}$ is the amount mutual information shared between $w_1$, $w_2$, $w_3$ and $w_4$,

$p(w_1w_2w_3w_4)$ is the probability of $w_1$, $w_{2,}$ $w_3$ and $w_4$ occurring together,

$p(w_1w_2\,w_3)\,*p(w_2w_3w_4)$ is the probability of occurrence of the trigram $w_1w_2w_3$ independent of the probability of the occurrence of the trigram $w_2w_3w_4$ in the corpus.

### *4.2.6.1.4.  Pent-gram NPMI*

$$I_{pent} = log_2 \frac{p(w_1w_2w_3w_4w_5)^2}{p(w_1w_2w_3w_4) * p(w_2w_3w_4w_5)}$$

Where, $I_{pent}$ is the amount mutual information shared between and $w_1$, $w_2$, $w_3$, $w_4$ and $w_5$

$p(w_1w_2\,w_3\,w_4\,w_5)$ is the probability of $w_1$, $w_{2,}$ $w_3$, $w_4$ and $w_5$ occurring together

$p(w_1w_2w_3w_4)\,*p(w_2w_3w_4w_5)$ is the probability of occurrence of the quad-gram $w_1w_2w_3w_4$ independent of the probability of the occurrence of the quad-gram $w_2w_3w_4w_5$ in the corpus.

### *4.2.6.1.5.  Observation Results for NPMI*

| English Corpus | Hindi Corpus | Bengali Corpus |
| --- | --- | --- |
| helter skelter | उथल पुथल | ট্র্যাভেল এজেন্সিতে |
| cintayatyeva mayi | ओत प्रोत | সারভাইভ্যাল ট্রেনিং |
| fait accompli | হট্টে কট্টে | আসপারগার সিন্ড্রোম |
| kirayat Swertia | তিতর বিতর | লোভ সংবরণ |

| English Corpus | Hindi Corpus | Bengali Corpus |
|---|---|---|
| laissez faire | फलते फूलते | মহাপীঠ তারাপীঠ |
| magnum opus | घिसी पिटी | নিওন সাইন |
| handspun handwoven | क्षत विक्षत | মর্মান্তিক দুর্ঘটনা |
| piper longum | भूरि भूरि | ইয়া মাসল |
| pather Panchali | हक्का बक्का | শুষে নেয়া |
| co operation | तड़क भड़क | যৌনতার আবেদনে |

**Table 4.2:** Observation results for Normalized PMI

Table 4.2 shows the top 10 candidates extracted by Normalized PMI method for English, Hindi and Bengali corpus.

### *4.2.6.2.  Bi-Directional Log-Likelihood Algorithm*

Log-Likelihood Ratio (LLR) is a hypothesis testing used to test whether the constituent words of an expression are dependent or independent of one another. Two hypotheses (of independence and dependence) for the occurrence frequency of a bigram (w1, w2) are considered:

Hypothesis 1: $P(w_2|w_1) = p = P(w_2|\neg w_1)$

Hypothesis 2: $P(w_2|w_1) = p_1 \neq p_2 = P(w_2|\neg w_1)$

Hypothesis 1 gives the likelihood of the occurrence of $w_2$ being independent of $w_1$, whereas hypothesis 2 measures the likelihood of the occurrence of $w_2$ being dependent on $w_1$. The ratio of these two likelihoods tells us, which hypothesis is more likely.

$$log_2(\lambda) = log_2 \frac{L(H_1)}{L(H_2)}$$

The issue with the state-of-the-art LLR is that it checks only for the dependence of $w_2$ on $w_1$ and does not consider the dependence (or independence) of $w_1$ on $w_2$. We modify the original formula to take into consideration dependence from both directions.

### *4.2.6.2.1.  Bi-gram Bi-directional Log-Likelihood Ratio*

Let $w_1w_2$ be the Bi-gram, the probabilities $p_1$, $p_2$, $p_3$ and $p_4$ are calculated as:

$$p_1 = p(w_2|w_1)$$

$$p_2 = p(w_2|\sim w_1)$$
$$p_3 = p(w_1|w_2)$$
$$p_4 = p(w_1|\sim w_2)$$

Hypothesis 1: $p_1 = p_2$      Hypothesis 2: $p_1 \neq p_2$

Hypothesis 3: $p_3 = p_4$      Hypothesis 4: $p_3 \neq p_4$

The Log-Likelihood score is calculated as:

$$Log_{bi} = (log_2\left(\frac{L(H_1)}{L(H_2)}\right) + log_2\left(\frac{L(H_3)}{L(H_4)}\right))/2$$

We expand the formulas of (n-1) gram to find out the score for n-gram.

### 4.2.6.2.2. Tri-gram Bi-directional Log-Likelihood Ratio

Let $w_1 w_2 w_3$ is the Tri-gram, the probabilities $p_1$, $p_2$, $p_3$ and $p_4$ are calculated as:

$$p_1 = p(w_3|w_1 w_2)$$
$$p_2 = p(w_3|\sim(w_1 w_2))$$
$$p_3 = p(w_1|w_2 w_3)$$
$$p_4 = p(w_1|\sim(w_2 w_3))$$

The Log-Likelihood score is calculated as:

$$Log_{tri} = (log_2\left(\frac{L(H_1)}{L(H_2)}\right) + log_2\left(\frac{L(H_3)}{L(H_4)}\right))/2$$

### 4.2.6.2.3. Quad-gram Bi-directional Log-Likelihood Ratio

Let $w_1 w_2 w_3 w_4$ is the Quad-gram, the probabilities are calculated as:

$$p_1 = p(w_4|w_1 w_2 w_3)$$
$$p_2 = p(w_4|\sim(w_1 w_2 w_3))$$
$$p_3 = p(w_1|w_2 w_3 w_4)$$
$$p_4 = p(w_1|\sim(w_2 w_3 w_4))$$

So the Log-Likelihood score is calculated as:

$$Log_{quad} = (log_2\left(\frac{L(H_1)}{L(H_2)}\right) + log_2\left(\frac{L(H_3)}{L(H_4)}\right))/2$$

### 4.2.6.2.4. Pent-gram Bi-directional Log-Likelihood Ratio

Let $w_1 w_2 w_3 w_4 w_5$ is the Pentagram, the probabilities $p_1$ and $p_2$ are calculated as:

$$p_1 = p(w_5|w_1w_2w_3w_4)$$
$$p_2 = p(w_5|\sim(w_1w_2w_3w_4))$$
$$p_3 = p(w_1|w_2w_3w_4w_5)$$
$$p_4 = p(w_1|\sim(w_2w_3w_4w_5))$$

The Log-Likelihood score is calculated as:

$$Log_{pent} = (log_2\left(\frac{L(H_1)}{L(H_2)}\right) + log_2\left(\frac{L(H_3)}{L(H_4)}\right))/2$$

### *4.2.6.2.5. Observation Results for Bi-directional Log-Likelihood Ratio*

| English Corpus | Hindi Corpus | Bengali Corpus |
|---|---|---|
| co operation | जा सकता | কথা বলিনি |
| supreme court | किया गया | করার পথনির্দেশিকা |
| cave temple | दिया गया | যৌন সঙ্গমে |
| nineteenth Century | पेज होम | হতে পারে |
| State governments | जाना चाहिए | ডাক্তার ল্যাপারোস্কোপির |
| rainy season | अगला पेज | চিন্তা উঠা |
| mother tongue | होना चाहिए | মর্মান্তিক দুর্ঘটনা |
| amendment Act | जा रहा | মানসিক চাপ |
| Folk Songs | किए गए | প্রাথমিক শিক্ষার |
| Information Technology | मंत्रि मण्डल | শারীরিক সুস্থতায় |

**Table 4.3:** Observation Results for Bi-directional LLR

Table 4.3 shows the top 10 candidates extracted by Bi-directional LLR method from English, Hindi, Bengali corpus.

### *4.2.6.3. Dice Coefficient*

The Sørensen–Dice index or Dice coefficient is a statistic used for comparing the similarity of two samples. We use dice coefficient for the detection of collocations in the following way,

$$Dice = \frac{c(w_1 w_2)}{c(w_1) + c(w_2)}$$

Where, c(w₁w₂) gives the count (frequency) of the bigram w₁w₂ in the corpus

c(w₁) is the count of w₁ and c(w₂) is the count of w₂ in the corpus

### 4.2.6.3.1. Dice Coefficient for Bigrams

For two words w₁ and w₂ in the corpus, their dice coefficient score is calculated as:

$$Dice_{Bi} = \frac{c(w_1 w_2)}{c(w_1) + c(w_2)}$$

Where, c(w₁w₂) gives the count (frequency) of the bigram w₁w₂ in the corpus

c(w₁) is the count of w₁ and c(w₂) is the count of w₂ in the corpus

### 4.2.6.3.2. Dice Coefficient for Tri-grams

For three words w₁, w₂ and w₃ in the corpus, their dice coefficient score is calculated as:

$$Dice_{Tri} = \frac{c(w_1 w_2 w_3)}{c(w_1 w_2) + c(w_2 w_3)}$$

Where, c(w₁w₂ w₃) gives the count (frequency) of the trigram w₁w₂w₃ in the corpus

c(w₁ w₂) is the count of bigram w₁ w₂ and c(w₂ w₃) is the count of bigram w₂w₃ in the corpus

### 4.2.6.3.3. Dice Coefficient for Quad-grams

For four words w₁, w₂, w₃ and w₄ in the corpus, their dice coefficient score is calculated as:

$$Dice_{Quad} = \frac{c(w_1 w_2 w_3 w_4)}{c(w_1 w_2 w_3) + c(w_2 w_3 w_4)}$$

Where, c(w₁w₂ w₃ w₄) gives the count (frequency) of the quad-gram w₁w₂w₃w₄ in the corpus

c(w₁ w₂ w₃) is the count of trigram w₁ w₂ w₃ and c(w₂ w₃ w₄) is the count of trigram w₂w₃ w₄ in the corpus

### 4.2.6.3.4. Dice Coefficient for Pent-grams

For five words w₁, w₂, w₃, w₄ and w₅ in the corpus, their dice coefficient score is calculated as:

$$Dice_{Pent} = \frac{c(w_1 w_2 w_3 w_4 w_5)}{c(w_1 w_2 w_3 w_4) + c(w_2 w_3 w_4 w_5)}$$

Where, c(w₁w₂ w₃ w₄ w₅) gives the count (frequency) of the pent-gram w₁w₂w₃w₄ w₅ in the corpus

c(w₁ w₂ w₃ w₄) is the count of quad-gram w₁ w₂ w₃ w₄ and c(w₂ w₃ w₄ w₅) is the count of quad-gram w₂w₃ w₄ w₅ in the corpus

### *4.2.6.3.5.* *Observation Results for Dice Coefficient*

| English Corpus | Hindi Corpus | Bengali Corpus |
|---|---|---|
| helter skelter | उथल पुथल | ট্র্যাভেল এজেন্সিতে |
| ekatvam anupasyatah | ओत प्रोत | বাসযাত্রা কমপ্লিমেন্টরি |
| magnum opus | हट्टे कट्टे | আসপারগার সিন্ড্রোম |
| handspun handwoven | तितर बितर | বান্দরপুণ পর্বতশ্রেণী |
| co operation | फलते फूलते | লোভ সংবরণ |
| pather Panchali | स्वै पूरणता | বোর্ডিং পাস |
| bullock cart | घिसी पिटी | ক্যান্সারের ঝুকি |
| suo moto | सिट्टी पिट्टी | অঙ্ক কষার |
| Homo erectus | क्षत विक्षत | উন্মেষ ঘটিল |
| pitter patter | हिले डुले | আত্মীয় স্বজনেরা |

**Table 4.4:** Observation Results for Dice Coefficient

Table 4.4 shows the top 10 candidates extracted by Dice coefficient method from English, Hindi, Bengali corpus.

### *4.2.6.4.* *Combining Different Filter's Scores*

To combine the scores of PMI, Log-Likelihood and Dice Coefficient, their independent scores are initially normalized. Normalization is done in the interval zero to one, by dividing score of each candidate by the maximum score obtained in the respective algorithm. After the normalization, all the three scores are added for each MWE candidate. After combining, the list is sorted to generate a combined ranked list.

### 4.2.7 Manual Evaluation

This is the final stage of the pipeline. The generated ranked list is evaluated by lexicographers to determine whether a candidate is truly MWE or not. The false positives are discarded by the lexicographers and the true MWEs are added to the dictionary. There is also an option for lexicographers to detect false negatives and add them to the dictionary.

The MWEs that are added to the dictionary are thus all validated by lexicographers and hence can serve as a gold standard.

### 4.2.8 Universal Web Service

Previously, the MWE engine is an offline Java application that needed to be installed locally and used. A multilingual web service has been developed in collaboration with Goa University to release an online interface to use the engine which will be hosted at IIT Bombay. This web service will be available to all research groups across India. Different language groups can upload their corpus and query the MWE extraction engine to extract MWE candidates from it. The results produced by the MWE extraction engine can later be validated by them.

In future, there will also be provision for the lexicographers to enter the true meaning of a non-compositional MWE so that these MWEs can be stored in lexical databases. Depending on the true meaning of an MWE the expression should be associated with the corresponding synset of Wordnet. By attaching gold standard MWEs to lexical resources we will not only be enhancing the resource but other applications will also be able to benefit from the knowledge of MWEs.

# Chapter 5

# Experimentations

## 5.1. Using Parallel Corpora

Multiword Expressions are idiosyncrasies of a language, or some conventional way of expressing things in a particular language. It is quite intriguing to see how multiword expressions behave across languages. Whether a multiword expression gets translated to a single word in another language, or does it have a literal translation of its constituents, or does it translate to its gloss in absence of such a concept in the other language or its corresponding expression in the other language also an MWE!

### 5.1.1 Multilingual Aspects of Multiword Expressions

We studied an English-Hindi parallel corpus and found the following types of transformations that multiwords undertake across languages:

|  | English Expression | Hindi Expression | Transliteration | Gloss |
| --- | --- | --- | --- | --- |
| Transliteration | social services | सोशल सर्विसेज | social services | an organized activity to improve the condition of disadvantaged people in society |
|  | Service record | सर्विस रेकॉर्ड | service record |  |
| Expansion | day care | पुरे दिन की देखभाल | pure din ki dekhbhal | childcare during the day while parents work |
| MWE -> Single Word | certificate | प्रमाण पत्र | praman patra | a document attesting to the truth of certain stated facts |
|  | function | काम काज | kaam kaj | the actions and activities assigned to or required or |

|  | English Expression | Hindi Expression | Transliteration | Gloss |
|---|---|---|---|---|
|  |  |  |  | expected of a person or group |
| MWE-> MWE | take action | कदम उठाना | kadam uthana | to do something |

**Table 5.1**: Multilinguality of Multiword Expressions

## 5.1.2 Motivation

Multiword expressions are abundant and typical usages of a language, which makes it very important to store such expressions in lexical resources. In order to store an expression in a lexical database we need both the expression and its meaning. Adding all the multiword expressions of a language manually to a resource is both expensive and time consuming. However extracting such expressions from corpus is challenging and 'automatically understanding' their meaning is fairly nontrivial.

The motivation behind using parallel corpora is to extract multiword expressions of a language with the help of another language and also retrieve the meaning of the expression from to parallel corpora. The idea is to use word alignment to detect corresponding expression for the mwe in other language and retrieve the meaning from.

The underlying assumption of alignment-based approaches to MWE extraction is that MWEs are aligned across languages in a way that differs from compositional expressions. Word alignment can be used in many ways for detecting multiword expressions:

- Sequences of length 2 or more in the source language that are aligned with sequences of length 1 or more in the target. The intuition comes from the fact that non –compositional multiword expressions are generally not translated literally across languages.
- Focus on misalignments: trust the quality of 1:1 alignments and search for MWEs exactly in the areas that word alignment failed to properly align. The reason for trusting misalignments is that, it is generally hard for an automatic word aligner to align multiword expressions, especially idioms.

## 5.1.3 Automatically adding MWEs to Wordnet

Our primary motivation behind using word alignment is to retrieve the meaning of the expression automatically and store them appropriately in a lexical resource. Consider the following example:

- दम तोड़ा (Hindi)
- Transliteration: dam toda
- Gloss: breath break
- Translation: die

'दम तोड़ा' is an idiom in Hindi, which means 'to die'. If we want to store this expression to wordnet we should store it in the synset of 'मरना (marna)' which also means 'to die'. In the parallel corpora if we have a pair of sentences, like,

- His grandfather died
- उसके दादाजी ने दम तोड़ा (uske dadaji ne dam toda)

Through word alignment if we are able to map 'दम तोड़ा' to 'died' in the sentence, then we can find the synset corresponding to 'die' in English Wordnet. There exists links between English and Hindi Wordnet synsets. The synset containing 'die' should be linked to the synset containing 'मरना (marna)' in Hindi Wordnet. Following this link we should be able to establish the association between 'मरना (marna)' and 'दम तोड़ा (dam toda)' and put them in the same synset.

Even though theoretically the above idea looks very intuitive, it is hard to achieve this computationally;

- **Difficulty in alignment:** We used GIZA++ word aligner to align the English-Hindi parallel corpora. The alignment quality was not satisfactory. Due to data sparsity problem in the corpus the aligner was producing a lot of misalignments.
- **Difficulty in detecting idioms:** Detecting idioms automatically by any extraction approach (statistical or linguistic) is not easy. Due to the poor performance of the automatic word aligner the word alignment methodology also could not be trusted to detect idioms.

However, in spite of the computational difficulties the above mentioned idea is novel and intuitive. With bigger corpus we might be able to overcome the data sparsity issue faced by word aligner and implement the algorithm for automatically adding mwe to lexical resources.

### 5.1.4 Case Study

In this section we present a case study that we performed during our investigation of the multilingual aspects of mwe. We present the steps that we followed, our observation and the inference that we drew from the experiment.

1) Consider a Hindi-English parallel corpora
2) Run MWE extraction engine (described in Chapter 4) on one side
3) Run automatic word aligner to word align parallel sentences
4) Retrieve the translations (marked by the word aligner) of the multiword expressions (extracted by MWE extraction engine)

We performed the above experiment by running MWE extraction engine on English side of the corpora and studied the Hindi translation equivalents of English 'collocations'. Following are the results from **top 100 collocations** in English:

|  | **English Collocation** | **Hindi Expression** | **Transliteration** |
|---|---|---|---|
| Collocation -> Single Word (20%) | south east | आग्नेय | agneya |
|  | folk songs | लोकगीत | lokgeet |
|  | bullock cart | बैलगाड़ी | byalgadi |
| Collocation -> Translation/ Transliteration (80%) | information technology | सूचना तकनीक | suchana taknik |
|  | state government | राज्य सरकार | rajya sarkar |
|  | amendment act | संशोधन अधिनियम | sanshodhan adhiniyam |
|  | world war | विश्व युद्ध | vishwa yuddha |
|  | text books | पाठ्य पुस्तक | pathya pustak |

|  | **English Collocation** | **Hindi Expression** | **Transliteration** |
|---|---|---|---|
|  | railway station | रेलवे स्टेशन | railway station |

**Table 5.2**: Multilinguality of Collocations

As shown in **Table 5.2**, among the top ranked **English collocations** 20% translate to a single word in Hindi and all of the rest 80% collocations are either transliterated or their components are literally translated; resulting in **collocations in Hindi**.

Does the above observation mean, collocations in one language are collocations in another language, i.e. **collocations are language independent**!

If the above postulate is true then mining collocations from one language can help us save the cost of extraction from other languages just by doing translations or transliteration when the concept is not lexicalized in the other language. We need to gather more multilingual evidences to establish this fact.

## 5.2. Investigating Sanskrit Traditions

Sanskrit being a very ancient and grammatically enriched language there exists many rules as to when and why words join together in order to form a new expression.

There exists a concept of 'samasa' where under certain conditions words join together and the resulting expression achieves an exocentric meaning i.e. a meaning completely different from its constituent words. We will discuss a few types of 'samasa' in this section where we can witness such rules in action.

### 5.2.1 बहु व्रीहिः समासः (Exocentric Compounds)

बहु व्रीहिः समासः (Bahubrihi Samasa) refers to a class of 'samasa' where compounds are formed by joining two separate words. This is similar to the case of non-compositional multiwords due to the fact that the meaning of the resulting expression lies outside the meaning of the constituent words. This is also termed as Exocentric compounds. Syntactically, बहु व्रीहिः समासः (Bahubrihi Samasa) are of two types:

- **समानाधिकरणबहुव्रीहि (samanadhikarana bahubrihi)**

  This category is also known as सामान्य (samanya) meaning 'regular'. Constituents of the compound belonging to this category undergo same case inflection in विग्रहवाक्यम् (Vigrahbakyam).

  - दत्तपशु (Sanskrit)

    **Transliteration:** dattapashu
    **Gloss:** the person who has received an animal

- **व्याधिकरणबहुव्रीहिः (vyadhikarana bahubrihi)**

  The compounds belonging to this subtype of Bahubrihi samasa have constituent words which undergo different inflectional variations while forming the compound.

  - चक्रपाणी (Sanskrit)

    **Transliteration:** chakrapani
    **Gloss:** The person who is having chakra in his hand

## 5.2.2 अव्ययीभावः समासः (Adverbial Compounds)

अव्ययीभावः समासः (Avyaibhav samasa) refers to the type of compound where most of the times the first member of the compound is an adverb and the whole compound functions as an adverb as well. Generally, compound belonging to this type do not have exocentric meaning. Syntactically, अव्ययीभावः समासः (Avyaibhav samasa) are of three types:

- **अव्ययपूर्वपदः (Avyaya Purbapada):** The first constituent of the compound is Avyaya

  Example:

  - उपकृष्णम् (Sanskrit)

    **Transliteration:** upakrishnam
    **Gloss:** near Krishna

- **अव्ययोत्तरपदः (Avyaya Uttarpada):** Avyaya appears as the second constituent of the compound.

  Example:
  - शाकप्रति (Sanskrit)

    **Transliteration:** shakprati

    **Gloss:** very small quantity of vegetable

- **अव्ययपदरहितः (Avyaya Padarahit):** None of the constituents in the compound is an avyay, yet the compound behaves like one. (तिष्ठद्गु प्रभृतीनि P2.1.17)

  Example:
  - तिष्ठद्गु (Sanskrit)

    **Transliteration:** tishthatgu

    **Gloss:** The time when the cow gives milk

## 5.2.3 तत्पुरुषः समासः (Determinative Compounds)

In a तत्पुरुषः समासः (tatpuruṣa samasa), the first component is in a case relationship with another and the meaning of the compound is mostly governed by the latter member. Meaning of the compound is endocentric i.e. compositional and can be inferred from the meaning of the constituents. The members of the compounds need to be in the same inflectional case i.e. प्रथमा (Prathama).

Example:
- नीलमेघ: (Sanskrit)

  **Transliteration:** nilmegh

  **Gloss:** blue cloud

- विद्याधनम (Sanskrit)

  **Transliteration:** vidyadhanam

  **Gloss:** knowledge is wealth

## 5.2.4 द्वन्द्व (Coordinating compounds)

A द्वन्द्व (dvandva) 'pair' or twin or Siamese compound refers to some concepts that could be connected in sense by the conjunction. Apart from Sanskrit, Dvandvas are common in some languages such as Chinese, Japanese, and some Modern Indic languages such as Hindi and Urdu, but less common in English[30].

द्वन्द्व (dvandva) compounds in Sanskrit can mostly be classifies into the following types:

- **इतरेतर द्वन्द्व (simple)**

    इतरेतर द्वन्द्व (Itaretara dvandva), is an enumerative compound word, the meaning of which refers to all its constituent members. The last member governs the gender and the inflections on the whole compound.
    Example:
    - रामलक्ष्मणौ (Sanskrit)

        **Transliteration:** Rama-Lakshmanau
        **Gloss:** Rama and Lakshmana
    - आचार्यशिष्यौ (Sanskrit)

        **Transliteration:** acharya-sishya
        **Gloss:** teacher and student

- **समाहार द्वन्द्व (collectives)**

    समाहार द्वन्द्व (Samahar dvandva) is a collective compound word. The meaning of the compound refers to the collection of its constituent members. The resultant compound word is in the singular number and is always neuter in gender.
    Example:
    - पाणिपादम् (Sanskrit)

        **Transliteration:** Pani-padam
        **Gloss:** hands and legs

We are exploring these different types of 'samasa' in order to relate them to Multiword Expressions in present day languages. Since Sanskrit is the mother of most of the Indian languages, the grammatical rules for many of them are derived from Sanskrit. We are trying to gain insight into why and how multiwords are formed. This insight should help us understand the problem more deeply and will also enable us to formulate some rules to solve it.

# Chapter 6

# Evaluation of MWE Extraction Engine

In this section we present the evaluation of the performance of different filters in detecting multiword expressions of varied types and characteristics. We have carried out our evaluation for the following languages,

- English
- Hindi
- Bengali

The details of the corpora are given in Table 6.1.

| Corpus | Size (No. of Words) |
|---|---|
| **English** | **4401696** |
| **Hindi** | **4605343** |
| **Bengali** | **181891** |

**Table 6.1**: Corpus Details

## 6.1. Evaluation of MWE engine on English Corpus

The size of the corpus is 4401696 words. We initially applied the Regular Expression filter on the English corpus to narrow down these words by only selecting patterns which can form multiword expressions. We skip the filters of reduplication, partial reduplication and complex predicates since these phenomenon are specific to Indian languages and are not observed in English. We apply the semantic filter based on Princeton Wordnet to detect semantic relationship and then statistical filters to detect collocations.

| Filter | Precision |
|---|---|
| **Semantic Filter** | 34% |
| **Hyphenation Filter** | 41% |
| **Statistical Filter** (for Bigrams) | 54.3% |

**Table 6.2**: Precision values of different filters for English

|  | Bi-directional Log-Likelihood Ratio | Normalized PMI | Dice Coefficient |
|---|---|---|---|
| **Bi-grams** | 70% | 34% | 59% |
| **Tri-grams** | 51% | 11% | 41% |
| **Quad-grams** | 20% | 10% | 41% |
| **Pent-grams** | 7% | 8% | 20% |

**Table 6.3**: Precision values of n-gram collocations

Table 6.2 shows the precision values of Semantic, Hyphenation and Statistical Filters. The statistical methods produce ranked lists of collocations. We have evaluated top 200 candidates for each method and have shown the average precision. A comparative evaluation of the three different statistical measures has been produced in Table 6.3, showing the variation of precision values as the candidate size increases. As we can see that Bi-directional log-likelihood ratio performs considerably better than other two methods for Bigrams, whereas Dice Coefficient looks more promising for higher n-grams.

## 6.2. Evaluation of MWE engine on Hindi Corpus

The Hindi corpus used for evaluation is parallel to the English corpus, whose evaluation was presented in the previous subsection, thus belong to the same domain and has 4605343 words. Initially a regular expression filter is applied to narrow down the candidates. Reduplication filter and Partial Reduplication Filter is applied followed by the Hyphenation filter and Semantic filter. Complex Predicate filter is applied to detect conjunct verbs. Finally, statistical measures are applied for detecting collocations.

| Filter | Precision |
|---|---|
| **Reduplication Filter** | 31% |
| **Partial Reduplication Filter** | 22% |
| **Semantic Filter** | 30% |
| **Hyphenation Filter** | 53% |
| **Complex Predicate Filter** | 87.2% |
| **Statistical Filter (for Bigrams)** | 26.3% |

**Table 6.4:** Precision values of different filters for Hindi

|  | Bi-directional Log-Likelihood Ratio | Normalized PMI | Dice Coefficient |
|---|---|---|---|
| **Bi-grams** | 10% | 32% | 37% |
| **Tri-grams** | 12% | 8% | 32% |

**Table 6.5:** Precision values for n-gram collocations

| Verb | # Hits | Correct Hits | Precision |
|---|---|---|---|
| होना (hona) | 38 | 26 | 68% |
| कर (kar) | 720 | 695 | 96.53% |
| देना (dena) | 134 | 131 | 97.76% |
| डालना (dalna) | 12 | 10 | 83.33% |
| उठाना (uthana) | 46 | 32 | 69.57% |
| लेना (lena) | 59 | 56 | 94.92% |
| उड़ (ud) | 8 | 8 | 100% |

**Table 6.6**: Precision Values of Conjunct Verbs

**Table 6.4** shows the precision values of Semantic, Hyphenation and Statistical Filters. The statistical methods produce ranked lists of collocations. We have evaluated top 200 candidates for each method and have shown the average precision. A comparative evaluation of the three different statistical measures has been produced in **Table 6.5**, showing the variation of precision values as the candidate size increases. **Table 6.6** shows the accuracy of identifying complex predicates with respect to different verbs.

As we can see, there are quite a few verbs which do not appear in the Hindi Verbalizer list but actually form conjuncts. These expressions would have gone undetected (false negatives) if we had only used the Verbalizer list. Among the statistical measures, we can observe that Dice Coefficient performs considerably better than other two methods.

## 6.3. Evaluation of MWE engine on Bengali Corpus

The Bengali corpus used for evaluation has 181891 words. Initially a regular expression filter is applied to narrow down the candidates. Reduplication filter and Partial Reduplication Filter is applied followed by the Hyphenation filter and Semantic filter. Finally, statistical measures are applied for detecting collocations.

| Filter | Precision |
|---|---|
| **Reduplication Filter** | **99.3%** |
| **Partial Reduplication Filter** | **38%** |
| **Hyphenation Filter** | **61.8%** |
| Statistical Filter (for Bigrams) | 41.3% |

**Table 6.7**: Precision values of different filters for Bengali

|  | Bi-directional Log-Likelihood Ratio | Normalized PMI | Dice Coefficient |
|---|---|---|---|
| **Bi-grams** | 46% | 36% | 42% |
| **Tri-grams** | 5% | 8% | 15% |

**Table 6.8**: Precision values for n-gram collocations

**Table 6.7** shows the precision values of Reduplication, Partial Reduplication, Semantic, Hyphenation and Statistical Filters. The statistical methods produce ranked lists of collocations. We have evaluated top 200 candidates for each method and have shown the average precision. A comparative evaluation of the three different statistical measures has been produced in **Table 6.8**, showing the variation of precision values as the candidate size increases.

As we can see that Bi-directional log-likelihood ratio performs marginally better than other two methods for Bigrams, whereas Dice Coefficient looks more promising for higher n-grams.

# Chapter 7

# Common Concept Hierarchy

We present IndoNet, a multilingual lexical knowledge base for Indian languages. It is a linked structure of wordnets of 18 different Indian languages, Universal Word dictionary and the Suggested Upper Merged Ontology (SUMO). This is a huge knowledge base and can serve as a very useful resource for multiword as well as other natural language processing tasks. The network also provides the necessary abstraction and standardization for storing multiword expressions. We discuss various benefits of the network and challenges involved in the development. The system is encoded in Lexical Markup Framework (LMF) and we propose modifications in LMF to accommodate Universal Word Dictionary and SUMO. This standardized version of lexical knowledge base of Indian Languages can now easily be linked to similar global resources.

## 7.1. Motivation

Lexical resources play an important role in natural language processing tasks. Past couple of decades has shown an immense growth in the development of lexical resources such as wordnet, Wikipedia, ontologies etc. These resources vary significantly in structure and representation formalism.

In order to develop applications that can make use of different resources, it is essential to link these heterogeneous resources and develop a common representation framework. However, the differences in encoding of knowledge and multilinguality are the major road blocks in development of such a framework. Particularly, in a multilingual country like India, information is available in many different languages. In order to exchange information across cultures and languages, it is essential to create architecture to share various lexical resources across languages.

In this report we present IndoNet[1], a lexical resource created by merging wordnets of 18 different Indian languages. These languages cover 3 different language families, Indo Aryan, Sino-Tebetian and Dravidian. Universal Word Dictionary[24] and an upper ontology, SUMO[15]

Universal Word (UW), defined by a headword and a set of restrictions which give an unambiguous representation of the concept, forms the vocabulary of Universal Networking Language. Suggested Upper Merged Ontology (SUMO) is the largest freely available ontology which is linked to the entire English WordNet [13]. Though UNL is a graph based representation and SUMO is a formal ontology, both provide language independent conceptualization. This makes [1]them suitable candidates for interlingua. IndoNet is encoded in Lexical Markup Framework (LMF), an ISO standard (ISO-24613) for encoding lexical resources[10].

The contribution of this work is twofold,
- We propose an architecture to link lexical resources of Indian languages.
- We propose modifications in Lexical Markup Framework to create a linked structure of multilingual lexical resources and ontology.

Though the architecture currently contains only Indian languages, it can easily be extended to support other languages.

## 7.2. Related Work

Over the years wordnet has emerged as the most widely used lexical resource. Though most of the wordnets are built by following the standards laid by English Wordnet [9], their conceptualizations differ because of the differences in lexicalization of concepts across languages. `Not only that, there exist "lexical gaps "where a word in one language has no correspondence in another language, but there are differences in the ways languages structure their words and concepts'. [16]

---

[1] Wordnets for Indian languages are developed in IndoWordNet project. Wordnets are available in following Indian languages: Assamese, Bodo, Bengali, English, Gujarati, Hindi, Kashmiri, Konkani, Kannada, Malayalam, Manipuri, Marathi, Nepali, Punjabi, Sanskrit, Tamil, Telugu and Urdu. These languages covers 3 different language families, Indo Aryan, Sino-Tebetian and Dravidian.
http://www.cfilt.iitb.ac.in/indowordnet

The challenge of constructing a unified multilingual resource was first addressed in EuroWordNet [27]. EuroWordNet linked wordnets of 8 different European languages through a common interlingual index (ILI). ILI consists of English synsets and serves as a pivot to link other wordnets. While ILI allows each language wordnet to preserve its semantic structure, it has two basic drawbacks as described in [26].

- An ILI tied to one specific language clearly reflects only the inventory of the language it is based on, and gaps show up when lexicons of different languages are mapped to it.
- The semantic space covered by a word in one language often overlaps only partially with a similar word in another language, resulting in less than perfect mappings.

Subsequently in KYOTO project, ontologies are preferred over ILI for linking of concepts of different languages. Ontologies provide language indpendent conceptualization, hence the linking remains unbiased to a particular language. Top level ontology SUMO is used to link common base concepts across languages. Because of the small size of the top level ontology, only a few wordnet synsets can be linked directly to the ontological concept and most of the synsets get linked through subsumption relation. This leads to a significant amount of information loss.

KYOTO project used Lexical Markup Framework (LMF) [10] as a representation language. `LMF provides a common model for the creation and use of lexical resources, to manage the exchange of data among these resources, and to enable the merging of a large number of individual electronic resources to form extensive global electronic resources[10].

WordNet-LMF was proposed to represent wordnets in LMF format [21]. It has further been modified to accommodate lexical relations[11]. LMF also provides extensions for multilingual lexicons and for linking external resources, such as ontology. However, LMF does not explicitly define standards to share a common ontology among multilingual lexicons.

Our work falls in line with EuroWordNet and Kyoto except for the following key differences,

- Instead of using ILI, we use a `common concept hierarchy' as a backbone to link lexicons of different languages.
- In addition to an upper ontology, a concept in common concept hierarchy is also linked to Universal Word Dictionary. Universal Word dictionary provides additional semantic information regarding argument types of verbs, that can be used to provide clues for selectional preference of a verb.

- We refine LMF to link external resources (e.g. ontologies) with multilingual lexicon and to represent Universal Word Dictionary.

## 7.3. IndoNet

IndoNet uses a common concept hierarchy to link various heterogeneous lexical resources. As shown in Figure 7.1, concepts of different wordnets, Universal Word Dictionary and Upper Ontology are merged to form the common concept hierarchy. Figure 7.1 shows how concepts of English WordNet (EWN), Hindi Wordnet (HWN), upper ontology (SUMO) and Universal Word Dictionary (UWD) are linked through common concept hierarchy (CCH). This section provides details of Common Concept Hierarchy and LMF encoding for different resources.

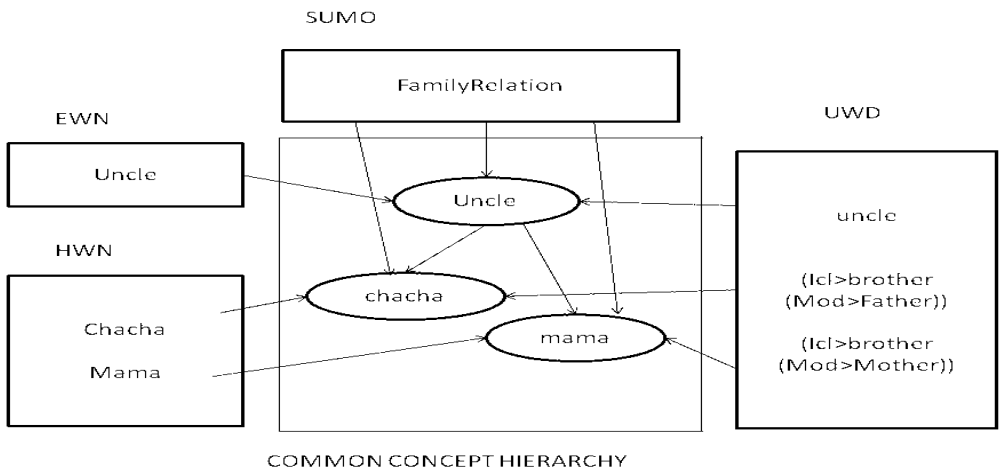

**Figure 7.1** : An Example of Indonet Structure

### 7.3.1 Common Concept Hierarchy (CCH)

The common concept hierarchy is an abstract pivot index to link lexical resources of all languages. An element of a common concept hierarchy is defined as <*sinid_1,sinid_2,..., uwid, sumoid*> where, *sinid_i* is synset id of $i^{th}$ wordnet, *uw_id* is universal word id, and *sumo_id* is SUMO term id of the concept. Unlike ILI, the hypernymy-hyponymy relations from different

Wordnets are merged to construct the concept hierarchy. Each synset of wordnet is directly linked to a concept in `common concept hierarchy'.

### 7.3.1.1. LMF for Wordnet

We have adapted the Wordnet-LMF, as specified in [21]. However IndoWordnet encodes more lexical relations compared to EuroWordnet. We enhanced the Wordnet-LMF to accommodate the following relations: *antonym, gradation, hypernymy, meronym, troponymy, entailment* and cross part of speech links for *ability* and *capability.*

### 7.3.1.2. LMF for Universal Word Dictionary

A Universal Word is composed of a headword and a list of restrictions, which provide unique meaning of the UW. In our architecture we allow each sense of a headword to have more than one set of restrictions (defined by different UW dictionaries) and are linked to lemmas of multiple languages with a confidence score. This allows us to merge multiple UW dictionaries and represent it in LMF format. We introduce four new LMF classes; *Restrictions, Restriction, Lemmas and Lemma* and add new attributes, *headword* and *mapping score* to existing LMF classes.

Figure 7.2 shows an example of LMF representation of UW Dictionary. At present, the dictionary is created by merging two dictionaries, UW++ [3] and CFILT Hin-UW. Lemmas from different languages are mapped to universal words and stored under the *Lemmas* class.

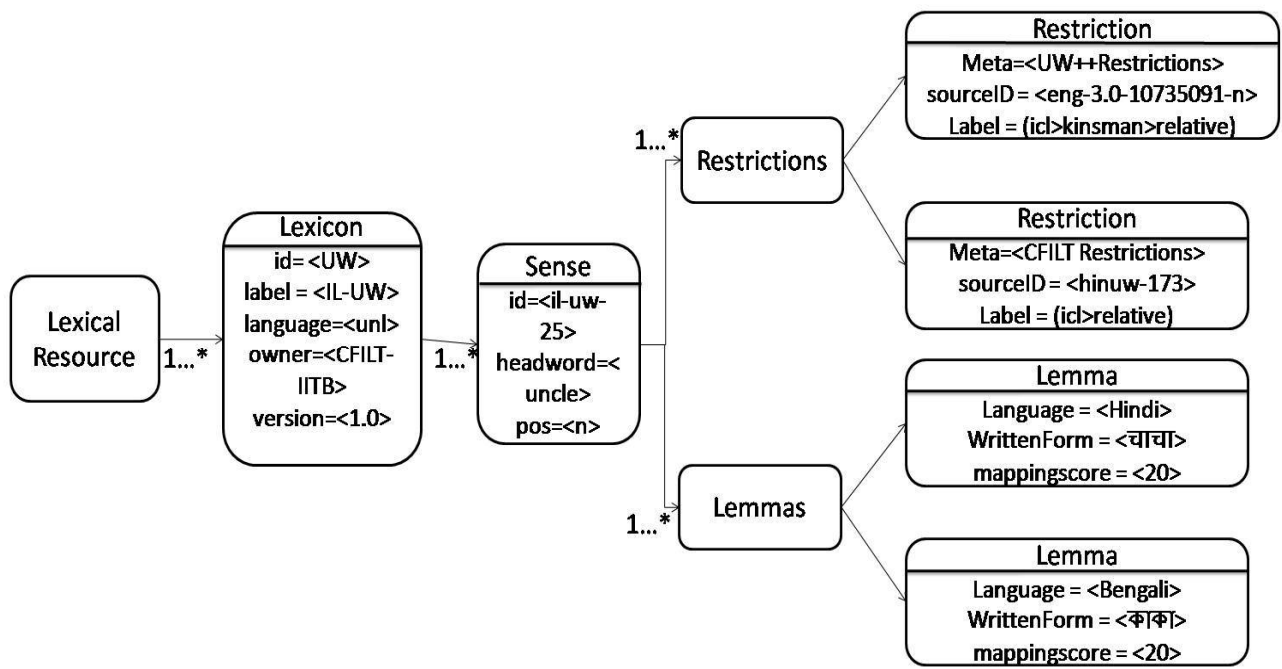

**Figure 7.2** : LMF representation for Universal Word Dictionary

### 7.3.1.3.    LMF to link ontology with Common Concept Hierarchy

Figure 7.3 shows an example LMF representation of CCH. The interlingual pivot is represented through *SenseAxis*. Concepts in different resources are linked to the *SenseAxis* in such a way that concepts linked to same *SenseAxis* convey the same *Sense*.

Using LMF class *MonolingualExternalRefs*, ontology can be integrated with a monolingual lexicon. In order to share an ontology among multilingual resources, we modify the original core package of LMF.

As shown in Figure 7.3, a SUMO term is shared across multiple lexicons via the *SenseAxis*. SUMO is linked with concept hierarchy using the following relations: *antonym, hypernym, instance and equivalent*. In order to support these relations, *Reltype* attribute is added to the interlingual *Sense* class.

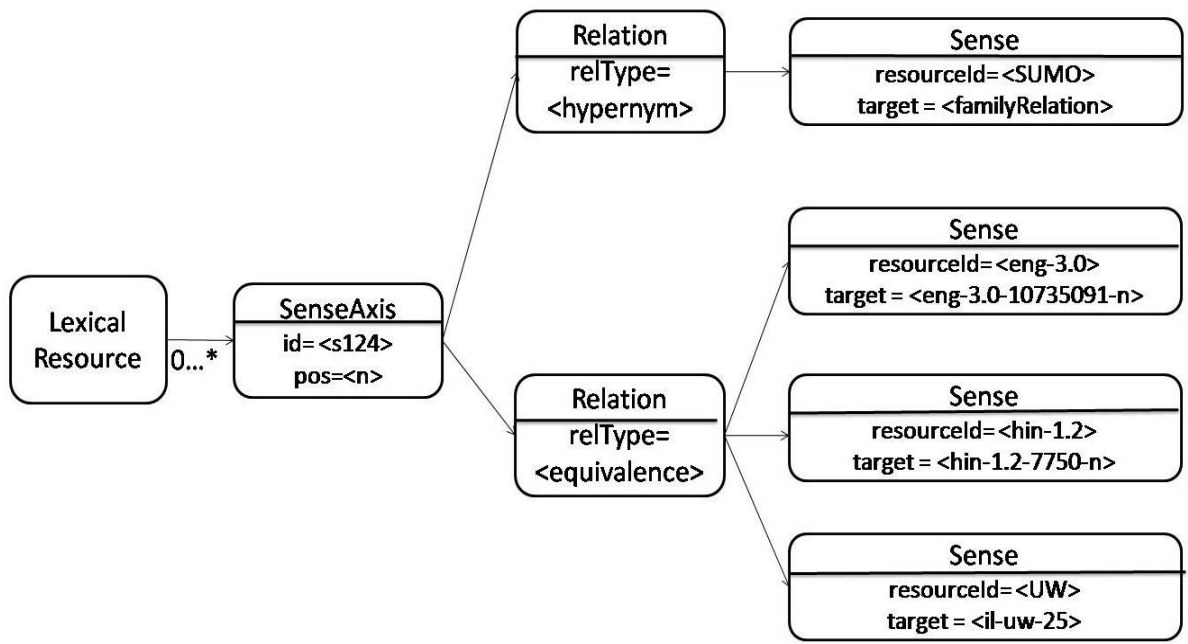

**Figure 7.3**: LMF representation for Common Concept Hierarchy

## 7.4. Observation

Table 7.1 shows *part of speech* wise status of linked concepts. The concept hierarchy contains 53848 concepts which are shared among wordnets of Indian languages, SUMO and Universal Word Dictionary. Out of the total 53848 concepts, 21984 are linked to SUMO, 34114 are linked to HWN and 44119 are linked to UW. Among these, 12,254 are common between UW and SUMO and 21984 are common between wordnet and SUMO.

| POS       | HWN   | UW    | SUMO  | CCH   |
|-----------|-------|-------|-------|-------|
| Adjective | 5532  | 2865  | 3140  | 5193  |
| Adverb    | 380   | 2697  | 249   | 2813  |
| Noun      | 25721 | 32831 | 16889 | 39620 |
| Verb      | 2481  | 5726  | 1706  | 6222  |
| Total     | 34114 | 44119 | 21984 | 53848 |

**Table 7.1**: Statistics of Concept Linkages

This creates a multilingual semantic lexicon that captures semantic relations between concepts of different languages. Figure 7.1 demonstrates this with an example of `kinship relation'. As shown in Figure 7.1, `*uncle'* is an English language concept defined as `the brother of your father or mother'. Hindi has no concept equivalent to `uncle' but there are two more specific concepts `*kaka'*, `brother of father.' and `*mama'*, `brother of mother.'

The lexical gap is captured when these concepts are linked to CCH. Through CCH, these concepts are linked to SUMO term `*FamilyRelation'* which shows relation between these concepts. Universal Word Dictionary captures exact relation between these concept by applying restrictions *[chacha] uncle(icl>brother (mod>father))* and *[mama] uncle(icl>brother (mod>mother))*. This makes it possible to link concepts across languages.

## 7.5. Conclusion

We have presented a multilingual lexical resource for Indian languages. The proposed architecture handles the `lexical gap' and `structural divergence' among languages, by building a common concept hierarchy. In order to encode this resource in LMF, we developed standards to represent UW in LMF.

IndoNet is emerging as the largest multilingual resource covering 18 languages of 3 different language families and it is possible to link or merge other standardized lexical resources with it. Since Universal Word dictionary is an integral part of the system, it can be used for UNL based Machine Translation tasks. Ontological structure of the system can be used for multilingual information retrieval and extraction.

In future, we aim to address ontological issues of the common concept hierarchy and integrate domain ontologies with the system. We are also aiming to develop standards to evaluate such multilingual resources and to validate axiomatic foundation of the same.

# Chapter 8

# Generic Stemmer

Stemmer is a basic building block for most of the natural language processing applications. Due to the morphological richness of Indian languages and lack of resources, stemmers are not available for many Indian languages. MWE extraction task required searching wordnet database for words read from a corpus, which might be in a morphed form. Wordnet doesn't store morphological variations of a word and hence we need to lemmatize it before searching wordnet. In this report, we present a generic stemmer that can be used for all major Indian languages based on IndoWordnet. The stemming algorithm is based on trie data structure and uses the lemmas from Wordnet to find out lexeme for a given word.

## 8.1. Construction of the Generic Stemmer

In this section we discuss the construction of the stemmer. IndoWordnet is a lexical database for 18 Indian languages. We use the words from wordnet as a dictionary to search for a lemma.

Given an input word we follow the following steps to search for its stem and lemma:
  i) Read the input word and input language
  ii) Search the word in wordnet of the appropriate language
  iii) If the word is found in the wordnet
      a. Output the word as the lemma as well as stem
      b. Exit
  iv) Find the string with maximal matching in Wordnet (maxMatch)
      a. Output maxMatch as the stem
      b. Find all words which have maxMatch as a prefix and output them as lemmas
  v) Ask the user if the output is correct
      a. If yes, then exit
      b. Else backtrack, remove the last character of maxMatch and go to step iv a.

### 8.1.1 Storing Wordnet in a Trie

Searching for words in Wordnet, requires an in-memory storage of the database for efficient performance. Since Wordnet is a huge lexical database containing thousands of words it'll be very inefficient to read the database files multiple times for each individual word. There can be multiple options for the data structure to store wordnet in-memory, we chose Trie.

### 8.1.2 Why Trie?

In computer science, a **trie**, also called **digital tree** or **prefix tree**, is an ordered tree data structure that is used to store a dynamic set where the keys are usually strings [23]. Unlike a binary search tree, no node in the tree stores the key associated with that node; instead, its position in the tree defines the key with which it is associated. All the descendants of a node have a common prefix of the string associated with that node, and the root is associated with the empty string. Values are normally not associated with every node, only with leaves and some inner nodes that correspond to keys of interest.

The term trie comes from re**trie**val. Trie is a very useful data structure for string retrieval. Looking up data in a trie is faster in the worst case, O(m) time (where m is the length of a search string), compared to an imperfect hash table. An imperfect hash table can have key collisions (a key collision is the hash function mapping of different keys to the same position in a hash table).

A common application of a trie is storing a predictive text or autocomplete dictionary. Such applications take advantage of a trie's ability to quickly search for, insert, and delete entries. Tries are also well suited for implementing approximate matching algorithms, including those used in spell checking and hyphenation software.

For our purpose, we need to store the lexical knowledge present in wordnet in a in-memory data structure which should provide prefix-matching and fast lookup. Hence trie was the obvious choice.

### 8.1.3 Structure of the Trie

The words from wordnet are stored in a trie data structure along with their part-of-speech tag and synset id.

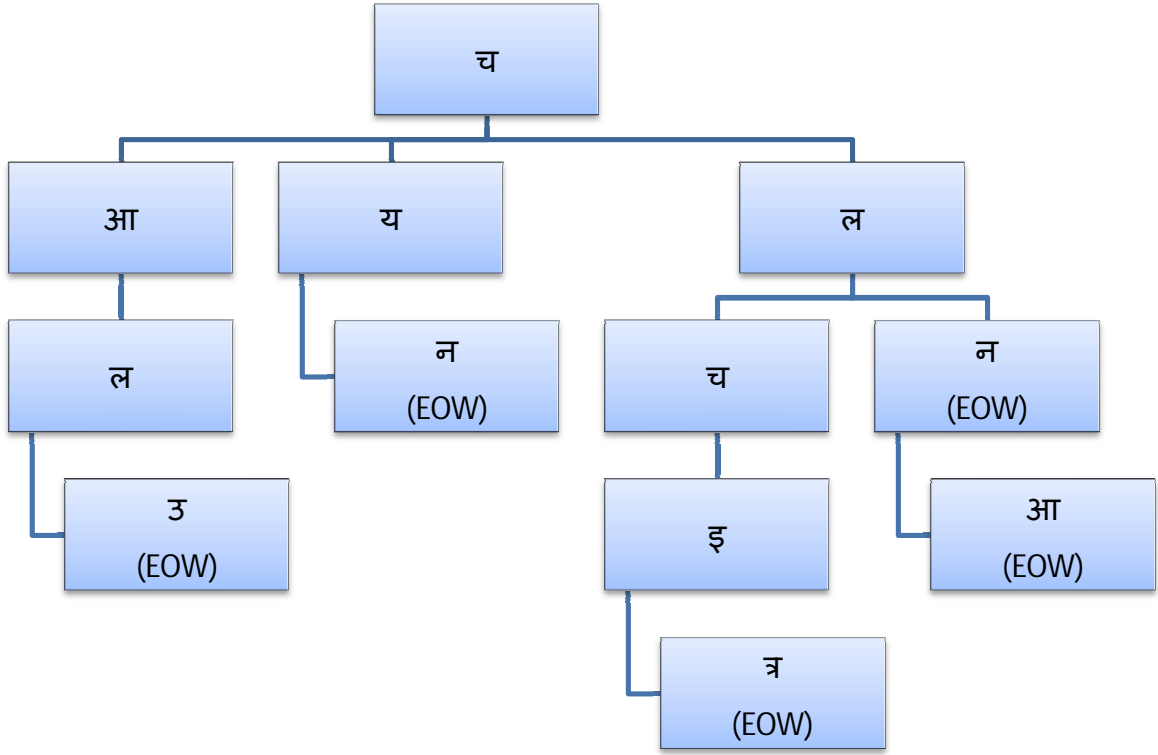

**Figure 8.1**: Example of Hindi wordnet representation in trie

Figure 8.1 shows an example to demonstrate how the words are stored in wordnet. Each leaf node contains a Boolean marker to denote 'end of word'. A non-leaf node might also be an 'end of word' as shown in figure. The nodes which have the EOW flag set to 'true', also stores the part-of-speech tag and synset-id(s) of the word.

For example, consider the input word to be चलती *(chalti)*. We start searching from the root of trie and try to match one character in each level. Consider the snippet of trie shown in Figure 8.1, we match the character 'च' and follow the pointer to the node 'ल'. After this node we can't find a node matching the next character, so this is the node of maximal matching. We store parent pointers at each node to trace back the path, hence after reaching this node; we trace the parent pointers back to the root node and return the string 'चल *(chal)*' as stem. In order to find the lemmas, we search all the children nodes of the maximal matching node, traverse each path till the leaf nodes and print the words whenever 'EOW' flag is encountered. For the input word 'चलती' if the algorithm is applied on the trie snippet shown in Figure 8.1 , 'चलना' and 'चलचित्र' will be returned as lemmas.

The problem with this approach is, it might return a lot of completely unrelated lemmas which just share the prefix with the input word. Some intelligent methodology can be applied to determine the ranking of the lemmas. We have applied a simple heuristic which ranks the lemmas according to its increasing difference of length with the root word.

The system provides a feedback mechanism for the user to convey whether the correct stem and lemma have been found. If it has not been found, the algorithm backtracks by going to the parent node of the maximal matching node and then producing stem and lemma from that node.

## 8.2. GUI of the Stemmer

In order to facilitate the easy use of stemmer it is provided with a graphical user interface developed using Java Swing API.

The system is integrated with IndoWordnet and hence can act as a generic stemmer catering to 18 Indian languages. The user interface allows a user to select the input language from a drop-down menu of all languages and enter an input word.

The system displays the stem and lemmas for the input word in different textboxes and asks the user whether the correct stem and lemma have been found. If the user thinks that the output is incorrect, he can provide his feedback through the radio buttons and the system backtracks and produces a new set of results by going one level up in the depth of the trie. This is a totally user driven system and the backtracking mechanism may continue till the system reaches root of the trie.

**Figure 8.2** shows a screenshot of the GUI of the generic stemmer. The input language is Hindi and input word is 'घुमते (ghumte)'. The system correctly finds the stem as 'घुम (ghum)' and identifies the lemma 'घुमना (ghumna)'.

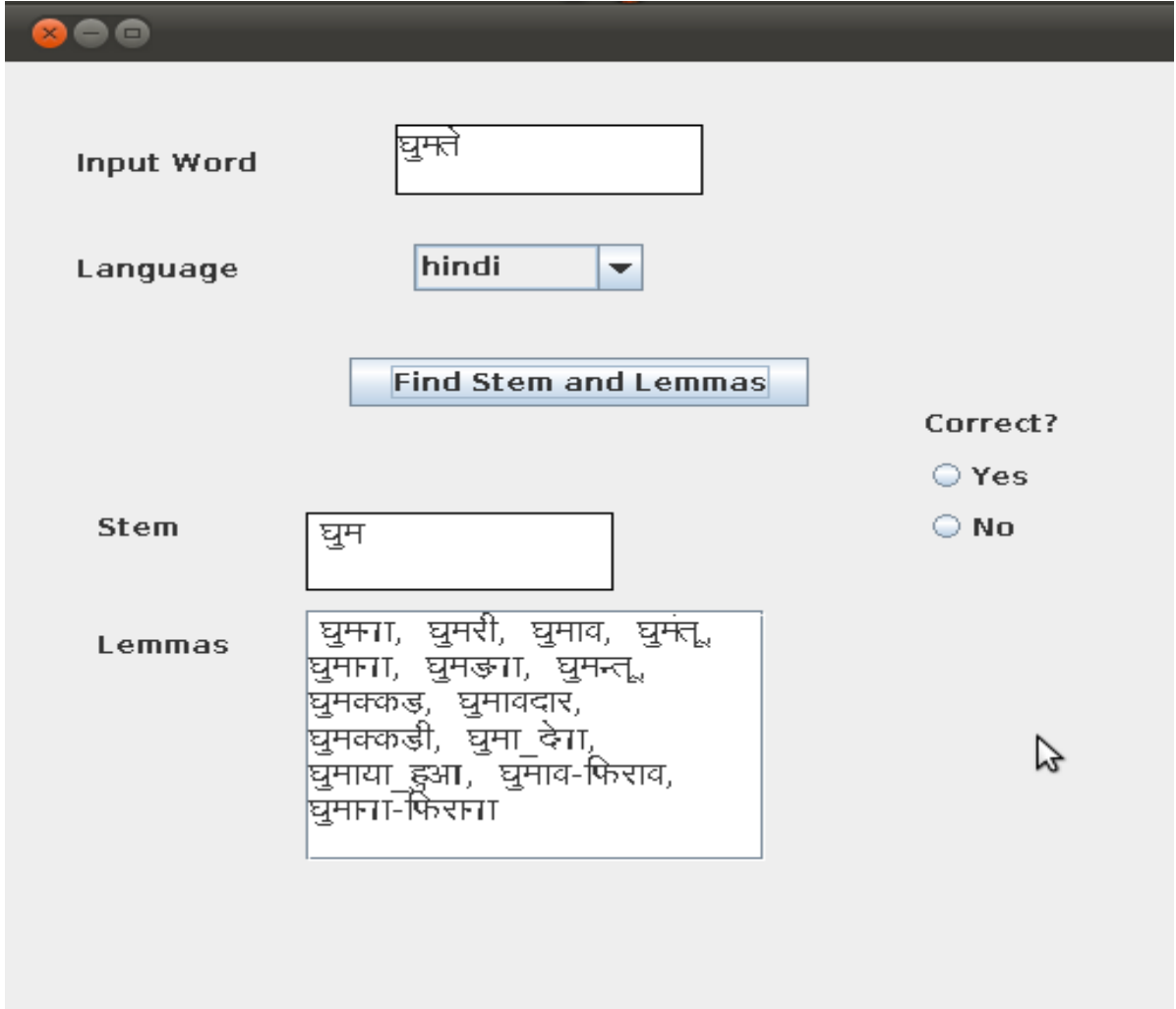

**Figure 8.2**: GUI of generic stemmer

## 8.3. Integration with Multilingual IndoWordnet Search

The generic stemmer that has been developed can not only serve as a stand-alone system, but can also be integrated with a number of applications to increase their efficiency and convenience of use. Since the whole wordnet is stored in a trie data structure, this library can serve as a spell checker or auto-completion tool for wordnet search very efficiently.

Indian languages are rich in morphology but morphological variations are not stored in wordnet. It is not be possible to store all the morphological variations of every word in wordnet but many applications may require searching a morphed word (for eg: an application that reads a corpus and uses wordnet for some analysis of the words).

There exists a multilingual web service for searching Indo Wordnet. We have integrated our system to the web service to provide automatic Word Suggestions. When a word is entered for searching, it is searched in the trie. If the word is present in Wordnet, a complete match of it will be found in trie and the page automatically redirects to show its synsets. However, if the word is not found, it searches for all the lemmas and provides them as 'Suggested Words' for the user to select. The backtracking option is provided with a link named 'Show More'. A screenshot of this is provided in **Figure 8.3**.

**Figure 8.3**: Multilingual IndoWordnet Search

# Chapter 9

# Conclusion and Future Work

## 9.1. Conclusion

Through this report we have established the notion of Multiword Expressions and the importance of them in all fields of Natural Language Processing.

We have presented a thorough literature survey in this paper. We have discussed various definitions of MWEs given by different researchers and their types and characteristics. We have also touched upon some of the distinguished MWE extraction approaches carried out by various researchers. There exist some ongoing project in research labs for MWE extraction, we have described one such system in detail.

We have presented the MWE extraction engine developed at IIT Bombay. It has a pipeline consisting of various filters to detect MWEs from a corpus. A web service is developed for MWE extraction with the engine operating at backend. The web service is on a completely multilingual platform where various language research groups in India can upload their corpus and extract MWEs. It will also help us to build Gold Standard MWE data which can be added to lexical resources.

We have also described a common concept hierarchy which can serve as a standardized knowledge network for multiword and many natural language processing tasks.

A language independent generic stemmer is developed as a byproduct of the multiword research, which also has wide applications. The construction of the stemmer as well as its integration with a multilingual wordnet search web service has been discussed.

## 9.2. Future Work

Future work of the project has been divided in three subsections depending on their complexity and importance to the project.

### 9.2.1 Short Term Future Work

**Adding MWEs to lexical resources**

As mentioned earlier, Multiword expressions need to be stored in lexical resources. For non-compositional multi-words it is extremely difficult (if not impossible) to detect the meaning automatically. While manual validation we need to store the meaning of such expressions, so that we can associate them with the corresponding synsets of Wordnet.

### 9.2.2 Medium Term Future Work

**Associating a chunker**

The definition of MWs does not bind them to be contiguous hence a multiword expression can constitute of two words which are separated by other words in between. In order to recognize such pattern we need some syntactic information such as phrase boundaries in the sentence. We plan to integrate a chunker along with the existing model to detect non-contiguous multiword expressions.

### 9.2.3 Long Term Future Work

**Building a Machine Learning model for classification**

Detecting MWEs can be looked upon as a classification problem solvable by a standard Machine Learning model, given a feature set and training data. In order to train such model we need some high quality Gold standard training data which is not available yet. We hope that with the web service we will be able to collect manually annotated data from all across the nation and hence build a Machine Learning model to associate with this pipeline.

# PUBLICATION

Brijesh Bhatt, Lahari Poddar, Pushpak Bhattacharyya; *IndoNet: A Multilingual Lexical Knowledge Network for Indian Languages*, Association for Computational Linguistics (**ACL**) **2013**, Sofia, Bulgaria, 4-9 August, 2013